\documentclass[10pt,twocolumn,letterpaper]{article}

\usepackage{cvpr}              

\usepackage{array}
\usepackage{varwidth} 

\usepackage{adjustbox}
\usepackage{csquotes}
\usepackage{epigraph}
\usepackage[most,skins,theorems]{tcolorbox}
\usepackage{colortbl}
\usepackage{float}
\usepackage{placeins}
\usepackage{stfloats}

\usepackage{multicol}
\usepackage{multirow}
\usepackage[super]{nth}

\usepackage{enumitem}
\usepackage{caption}
\usepackage{wrapfig}

\definecolor{lightblue}{rgb}{0.941, 0.973, 1.0}
\definecolor{cvprblue}{rgb}{0.21,0.49,0.74}

\definecolor{cvprblue}{rgb}{0.21,0.49,0.74}
\usepackage[pagebackref,breaklinks,colorlinks,allcolors=cvprblue]{hyperref}

\def\paperID{} 
\def\confName{CVPR}
\def\confYear{2026}

\newcommand{\name}{VISOR}

\title{VISion On Request: Enhanced VLLM efficiency with sparse, dynamically selected, vision-language
interactions}

\author{
    Adrian Bulat$^{1,2}$\thanks{Equal contribution} \quad 
    Alberto Baldrati$^{1*}$ \quad 
    Ioannis Maniadis Metaxas$^{1*}$ \quad 
    \\
    Yassine Ouali$^1$ \quad 
    Georgios Tzimiropoulos$^{1,3}$ \\
    $^1$Samsung AI Cambridge\quad $^2$Technical University of Iasi \quad $^3$Queen Mary University of London
}

\begin{document}
\maketitle

\begin{abstract}

Existing approaches for improving the efficiency of Large Vision-Language Models (LVLMs) are largely based on the concept of visual token reduction. This approach, however, creates an information bottleneck that impairs performance, especially on challenging tasks that require fine-grained understanding and reasoning. In this work, we challenge this paradigm by introducing VISion On Request (\name), a method that reduces inference cost without discarding visual information. Instead of compressing the image, \name{} improves efficiency by sparsifying the interaction between image and text tokens. Specifically, the language model attends to the full set of high-resolution visual tokens through a small, strategically placed set of attention layers: general visual context is provided by efficient cross-attention between text-image, while a few well-placed and dynamically selected self-attention layers refine the visual representations themselves, enabling complex, high-resolution reasoning when needed. 
Based on this principle, we first train a single universal network on a range of computational budgets by varying the number of self-attention layers, and then introduce a lightweight policy mechanism that dynamically allocates visual computation based on per-sample complexity. Extensive experiments show that \name{} drastically reduces computational cost while matching or exceeding state-of-the-art results across a diverse suite of benchmarks, and excels in challenging tasks that require detailed visual understanding.

\end{abstract}

\section{Introduction}
Large Vision-Language Models (LVLMs) have demonstrated remarkable capabilities in multimodal understanding~\citep{yang2024qwen2,chen2024internvl,li2024llavaone}. These systems typically pair a vision encoder (e.g., CLIP~\citep{radford2021learning}) with a large language model (LLM)~\citep{touvron2023llama,jiang2023mistral,yang2024qwen2}. The vision encoder maps an input image to dense visual tokens, which are passed through a connector module and fed to the LLM alongside the textual prompt/query. Most of LVLM's computations are due to the large number of visual tokens, a cost that increases sharply with image resolution~\citep{li2024llavaone}. 
To mitigate this, a large volume of work has been proposed that explores the idea of \textit{token reduction/compression.} These works reduce the number of visual tokens by dynamically pruning and/or merging redundant tokens at test-time~\citep{arif2025hired,xing2025pyramiddrop,yang2025visionzip,zhang2024sparsevlm,chen2024image} or by training specialised compressors~\citep{cai2025matryoshka,hu2024matryoshka,chu2024mobilevlm}. While they perform well on tasks requiring coarse visual understanding, we show that they often incur substantial information loss on complex, high-resolution tasks that require fine-grained visual understanding. See accuracy on ``easy'' vs ``hard'' in Fig.~\ref{fig:efficiency-plot}. This is not surprising as such approaches, by shrinking the set of visual tokens, inevitably, create an information bottleneck. 

In this work, we propose a completely different and orthogonal path to token compression methods for increasing the efficiency of LVLMs. Unlike prior token reduction/compression methods that aim to reduce the number of visual tokens processed by the LVLM, our approach aims to reduce/sparsify the number of computational layers executed within the LVLM. Specifically, our method strategically executes a limited number of cross-attention and self-attention layers within the LVLM, allowing it to attend and update the full set of visual tokens only at a few selected points during the forward pass. Owing to this property, we coin our method - \name, Vision on Request. Our idea builds upon the observation that the query and answer tokens sparsely interact with the visual tokens~\citep{kaduri2025s} on a select few \textit{critical} layers. A phenomenon that we show to be heavily task-dependent, with the location and number of layers and the degree of sparsity varying significantly across tasks, depending on those tasks' complexity.

Overall, \textbf{we make the following contributions}: 
\begin{itemize}[leftmargin=*]
\itemsep0em 
    \item \textbf{Firstly}, we decompose the LVLM layer into image-image and text-image (cross-modal) interactions, and show that executing a fairly small number of cheap cross-attention layers for text-image, which operate on \textit{the same vision representations}, suffices for tasks requiring coarse visual understanding. This alone surpasses prior state-of-the-art methods on a range of vision-language benchmarks in terms of accuracy and speed.
    \item 
    \textbf{Secondly}, we demonstrate that for complex tasks, both prior works and our cross-attention only variant struggle to perform fine-grained visual understanding. We attribute this to the fact that cross-attention layers enable language tokens to attend to image information, but do not update/modify the visual tokens themselves. To alleviate this, we introduce and execute a small number of self-attention layers that perform the \textit{update of the visual tokens}, enabling a gradual refinement from lower to higher-level visual features.
    \item 
     \textbf{Thirdly}, as different tasks and samples require different amounts of visual detail, we first train a single universal network on a range of computational budgets by varying the number of self-attention layers. Then, we propose an \textit{adaptive inference} approach that automatically selects the self-attention layers to be executed on a per-sample basis using a lightweight policy mechanism trained via offline pseudo-labelling.
     \item 
     \textbf{Fourth}, we show that \name~can be \textit{combined with existing token reduction} methods to further improve efficiency without compromising performance.
     \item 
     \textbf{Fifth}, we set a new state-of-the-art on a range of vision language benchmarks, excelling in challenging tasks that require detailed visual understanding. See Fig.~\ref{fig:efficiency-plot}.
\end{itemize}

\section{Closely related work}

\noindent \textbf{Efficient LVLMs via Token Reduction:} To address the computational challenges posed by the large number of visual tokens in LVLMs, several approaches have been proposed to reduce the number of tokens processed by the LLM. These methods can be broadly grouped into two categories: dynamic token pruning and merging techniques~\citep{zhang2024sparsevlm,zhang2024cls,xing2025pyramiddrop,arif2025hired,shang2024llavaprumerge}, and learned token compression strategies~\citep{chu2024mobilevlm,yang2025visionzip,cai2025matryoshka,hu2024matryoshka,bulat2025compress}. 
\begin{figure}
    \vspace{-0.2cm}
    \centering
    \includegraphics[width=0.45\textwidth]{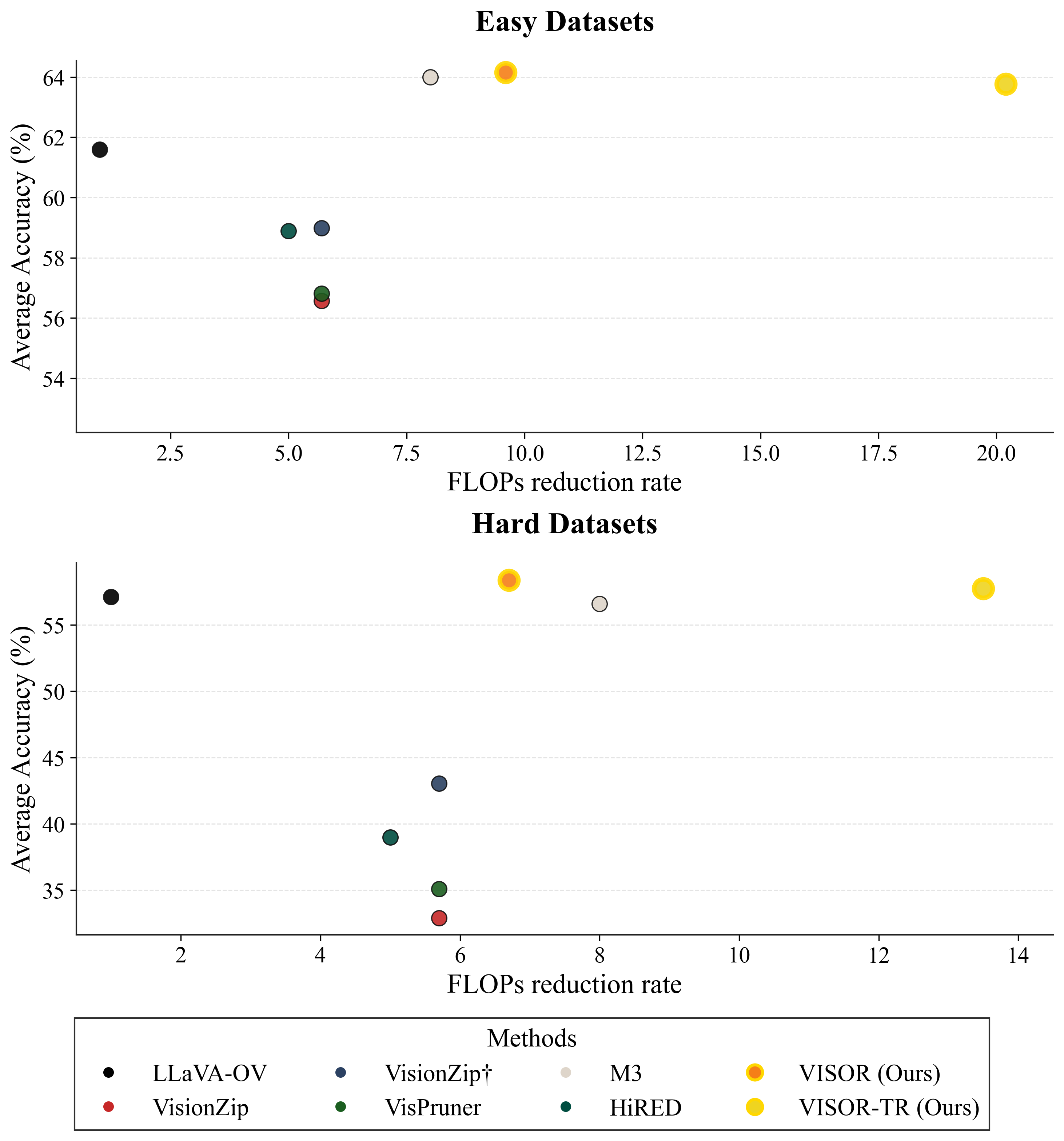}
    \vspace{-0.3cm}
    \caption{\textbf{Efficiency comparison:} FLOPs reduction vs acc. Notice that our approach is significantly more efficient while also retaining the performance on the harder datasets. See Sects.~\ref{sec:motivation} and ~\ref{ssec:experiments-setup} for ``easy’’-``hard’’ definition.}
    \label{fig:efficiency-plot}
    \vspace{-0.5cm}
\end{figure} 
The former category focuses on dynamically identifying the most important tokens, reducing redundancy by pruning or merging the less relevant tokens prior to the LLM~\citep{yang2025visionzip,zhang2024cls,shang2024llavaprumerge}, layer-by-layer within the LLM~\citep{xing2025pyramiddrop,chen2024image,yang2025topv,tan2025tokencarve}, or both~\citep{zhang2025vscan}, using heuristic criteria. Examples of criteria include: selecting top-k attended tokens~\citep{chen2024image}, assessing the correlation between patches~\citep{zhang2024token}, using the attention score between image tokens and [CLS] token~\citep{zhang2024cls}, rating the vision tokens using the text tokens~\citep{zhang2024sparsevlm}, or by analysing the information quantity in the attention matrix~\citep{tan2025tokencarve}. 
The latter category either replaces the connector module with a learned compressor~\citep{chu2024mobilevlm}, or introduces a new module before the LLM~\citep{yang2025visionzip} or as part of the vision encoder~\citep{cai2025matryoshka,hu2024matryoshka}. These methods finetune the LVLM, either fully or partially.

While showing promising results, most of these approaches focus on coarser understanding and lower-resolution tasks, often using a LLaVA-1.5 model~\citep{cai2025matryoshka,yang2025visionzip,bulat2025compress}. Very few works (e.g.,~\citep{arif2025hired,lan2025avg}) consider more challenging and fine-grained tasks that require higher resolution, with those that do either suffering from a large accuracy drop~\citep{arif2025hired} or exhibiting little to no speed-ups on these datasets~\citep{lan2025avg}. In this work, we further evaluate existing methods under a unified setting and architecture, and highlight this as a general trend in existing token reduction works. We argue that this performance degradation stems primarily from the information bottleneck inherent in token reduction. 

To alleviate this, \name{} sidesteps the token reduction paradigm altogether. Instead of reducing cost by discarding tokens, \name~ strategically limits the number of layers where the language model interacts with and updates visual information, thereby maintaining access to the full, high-resolution visual context throughout the model. This ensures that critical visual details are never permanently lost and can be accessed by the model when needed for fine-grained reasoning, while still achieving significant computational savings. Furthermore, our approach is orthogonal to existing token compression methods and can be combined with them for further efficiency gains.

\section{Motivation: Image processing within LVLMs}~\label{sec:motivation}

To motivate our design, herein, we focus on the internal workings of a standard LVLM (LLaVA-OV) to understand how it utilizes and processes visual information. We analyze the attention patterns of image-image and text-image (cross-modal) interactions and investigate three key questions:

\begin{figure*}[!ht]
    \centering
    \begin{subfigure}{0.32\linewidth}
        \centering
        \includegraphics[width=0.85\linewidth]{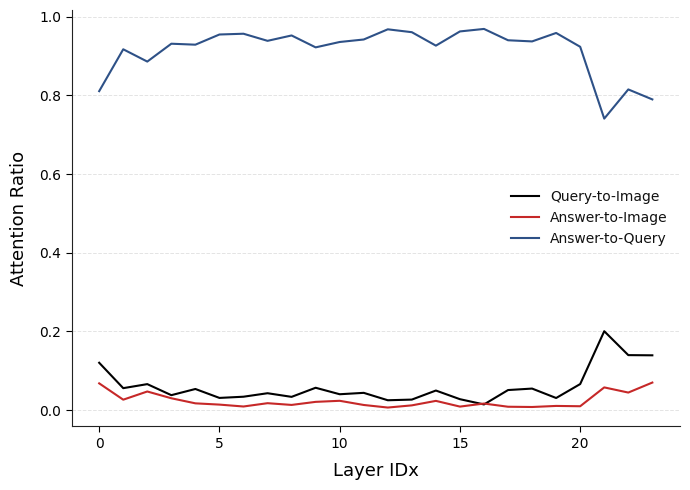}
        \caption{SQA}
        \label{fig:attn-sqa}
    \end{subfigure}
    \hfill
    \begin{subfigure}{0.32\linewidth}
        \centering
        \includegraphics[width=0.85\linewidth]{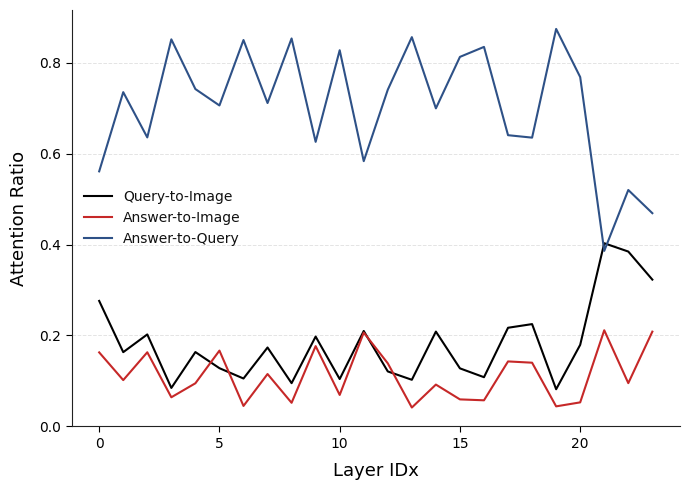}
        \caption{GQA}
        \label{fig:attn-gqa}
    \end{subfigure}
    \hfill
    \begin{subfigure}{0.32\linewidth}
        \centering
        \includegraphics[width=0.85\linewidth]{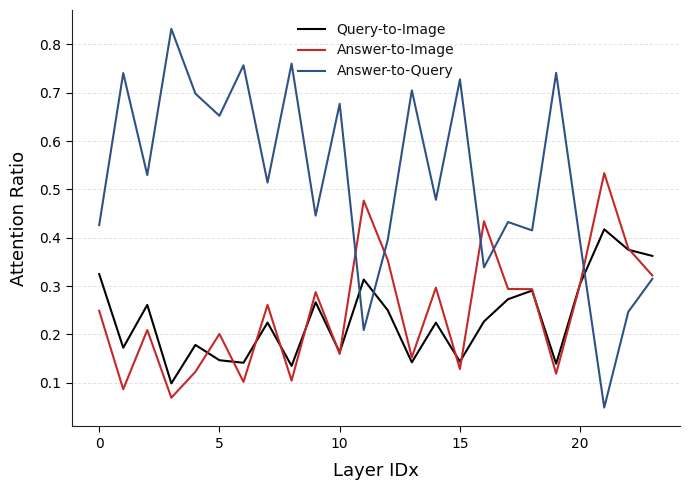}
        \caption{DocVQA}
        \label{fig:attn-doc-vqa}
    \end{subfigure}
    \vspace{-0.3cm}
    \caption{\textbf{Cross-modality attention patterns across layers.} We plot the proportion of attention scores allocated to three interaction types: text queries attending to image tokens (Query-to-Image), answer tokens attending to image tokens (Answer-to-Image), and answer tokens attending to query tokens (Answer-to-Query). For easy tasks like SQA, interaction is sparse and dominated by text-to-text attention. For hard tasks like DocVQA, the model attends to the image across the whole network.}
    \label{fig:attention-maps}
\end{figure*}

\textbf{How often, and when, does the model look at the image?} We distinguish between three types of interactions: Query-to-Image, Answer-To-Image, and Answer-To-Query. Fig.~\ref{fig:attention-maps} shows the layer-wise distribution of these interactions for three representative datasets. The results reveal that image-text interactions are task-dependent. 
For tasks requiring coarse vision understanding (e.g., ScienceQA), the model relies heavily on textual context (Answer-to-Query), with only limited interaction with the image, primarily in the initial and final layers. In contrast, for fine-grained tasks (e.g., DocVQA), the model exhibits sustained attention to the image across the whole network, indicating a continuous need for visual grounding. Moreover, we can observe that critical text-image interactions also occur in the middle layers in addition to the first and last layers. Interestingly, the saw-tooth patterns (for both GQA and DocVQA) suggest that not all cross-attention layers are necessary.

\begin{figure*}[!ht]
    \centering
    \begin{subfigure}{0.32\linewidth}
        \centering
        \includegraphics[width=0.8\linewidth]{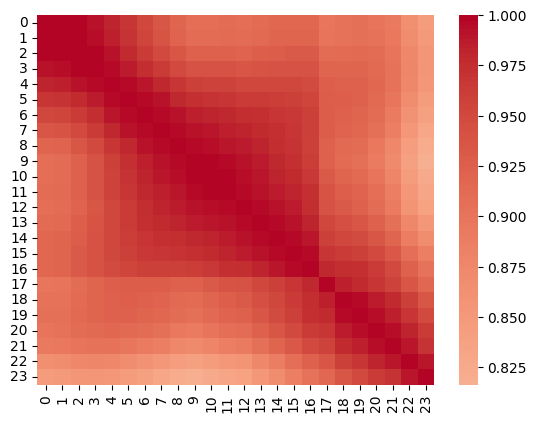}
        \caption{SQA}
        \label{fig:attn-sqa-feats-sim}
    \end{subfigure}
    \hfill
    \begin{subfigure}{0.32\linewidth}
        \centering
        \includegraphics[width=0.8\linewidth]{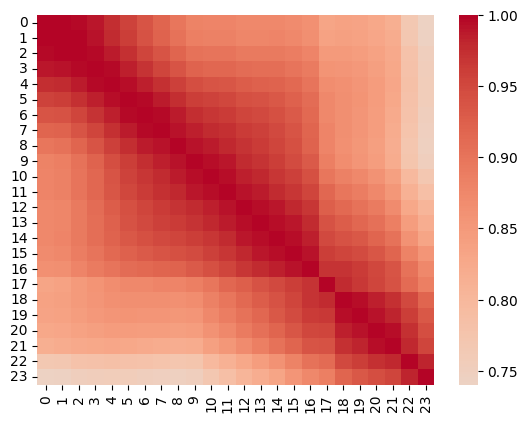}
        \caption{GQA}
        \label{fig:attn-gqa-feats-sim}
    \end{subfigure}
    \hfill
    \begin{subfigure}{0.32\linewidth}
        \centering
        \includegraphics[width=0.8\linewidth]{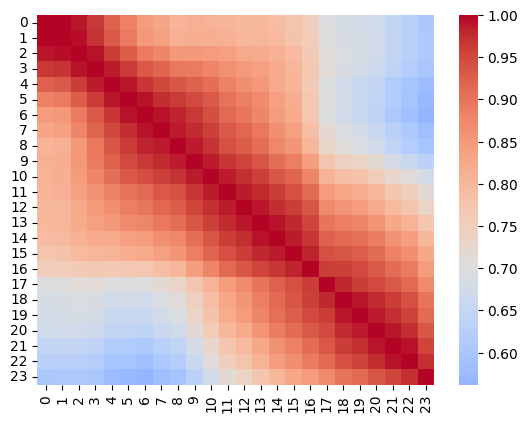}
        \caption{DocVQA}
        \label{fig:doc-vqa-feats-sim}
    \end{subfigure}
    \vspace{-0.3cm}
    \caption{\textbf{Evolution of visual representations across layers}, measured by pairwise CKA similarity. For easy tasks (e.g., SQA), visual features remain largely static (high similarity across layers). For harder tasks (e.g., DocVQA), features are progressively refined.}
    \label{fig:feats-similarity}
    \vspace{-0.3cm}
\end{figure*}

\textbf{How do visual representations evolve?} To analyze how vision features evolve across layers within the LLM transformer of the LVLM, we adopt the Centered Kernel Alignment (CKA)~\citep{cortes2012algorithms} similarity metric following~\citet{kornblith2019similarity} and~\citet{raghu2021vision} (see also supplementary material). 

We compute the pairwise CKA similarity between vision features from all layers of LLaVA-OV transformer on three representative datasets. As shown in Fig.~\ref{fig:feats-similarity}, for easy tasks like ScienceQA, the visual features remain largely unchanged throughout the model (CKA $>$ 0.9), implying that the initial representations are sufficient. However, for hard tasks like DocVQA, the features evolve significantly (CKA drops to 0.6), indicating that the model actively refines visual representations to solve the task. This highlights that while coarse tasks can rely on static visual features, complex tasks benefit from the refinement of visual information within the LLM. 
From the figure, we also observe a series of clusters emerging, indicating that the model refines visual features in stages. The number of stages is task-dependent, and we posit that it indicates the minimum number of self-attention layers that need to be executed to achieve optimal performance.

\begin{figure*}[!ht]
    \centering
    \begin{subfigure}{0.54\linewidth}
        \centering
        \includegraphics[trim={0 0 0 0.57cm},clip,width=\linewidth]{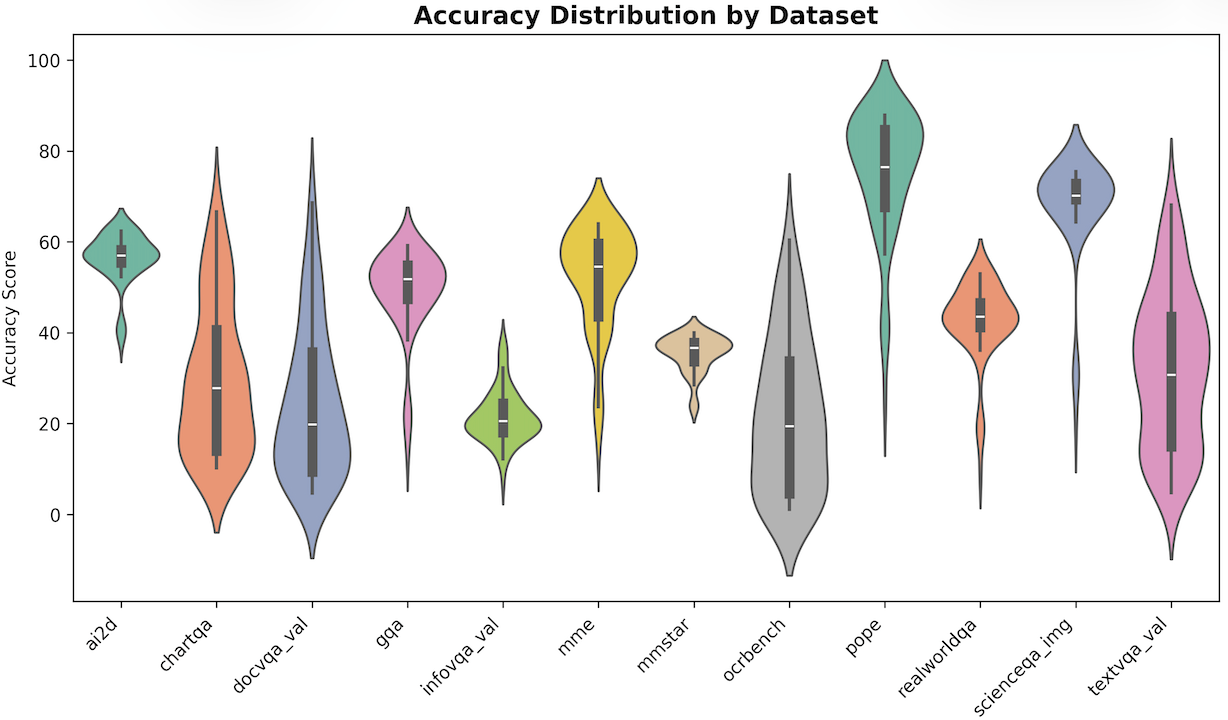}
         \caption{Accuracy distribution per dataset.}
        \label{fig:perf-dist}
    \end{subfigure}
    \hfill
    \begin{subfigure}{0.36\linewidth}
        \centering
        \includegraphics[trim={0 0 0 0.55cm},clip,width=\linewidth]{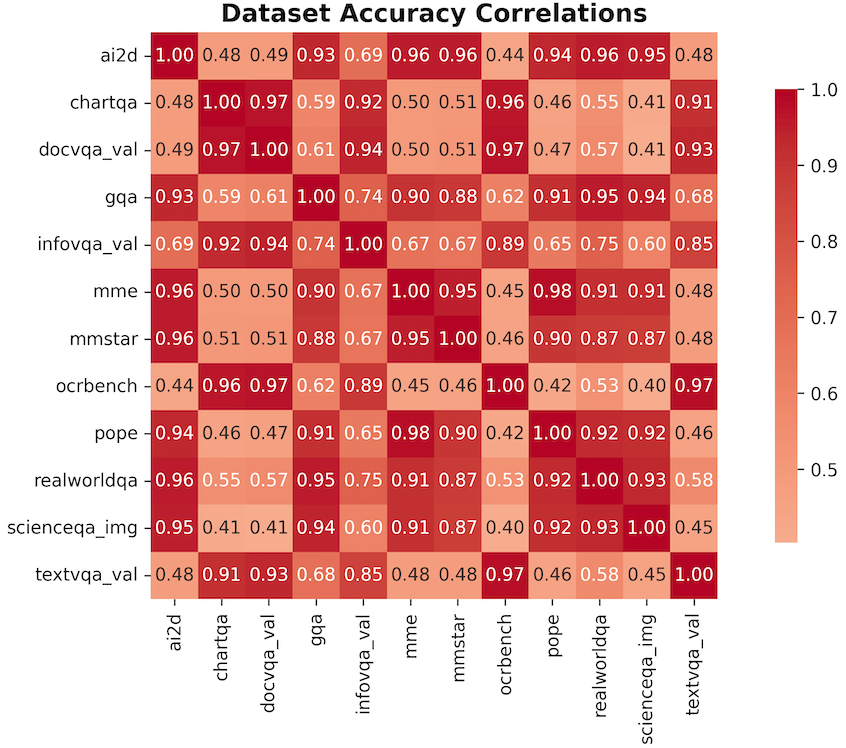}
        \caption{Dataset accuracy correlation.}
        \label{fig:dataset-perf-correlation}
    \end{subfigure}
    \vspace{-0.3cm}
    \caption{\textbf{Accuracy sensitivity by dropping all vision tokens for different subsets of LLM layers.} Left: Accuracy distribution on a dataset-by-dataset basis. Certain datasets (e.g., DocVQA, ChartQA) are particularly sensitive to reduced vision-language interactions. Right: we show how the layer-drop config. \& accuracy correlate among datasets. Two clusters emerge: vision-sensitive (``hard'') (e.g., InfoVQA, OCRBench, etc.) and coarse vision (``easy'') (e.g., POPE, SQA, GQA, etc.) datasets.
    }
    \label{fig:perf-layer-drop}
    \vspace{-0.3cm}
\end{figure*}

\textbf{What is the impact of reducing image-text interactions?} To this end, we drop \textit{all the vision tokens} from random subsets of LLM layers during inference and measure the performance degradation. Fig.~\ref{fig:perf-layer-drop} shows that datasets cluster into two groups. ``Easy'' tasks (e.g., SQA, POPE) are robust to this dropout, maintaining high performance. ``Hard'' tasks (e.g., DocVQA, ChartQA, InfoVQA) are highly sensitive, with performance dropping sharply as visual processing is reduced. We use this as a basis for dataset categorization in the rest of the paper. This confirms that a one-size-fits-all approach to visual processing is suboptimal; the computational budget should adapt to the sample/task at hand.

\textbf{Key takeaways} that inform the design of our proposed method: (1) Image-text interactions are sparse, exhibit saw-tooth patterns, and the degree of interaction is highly task-dependent. (2) While coarse tasks can rely on static visual features, complex tasks benefit from dynamic refinement of visual information within the LLM. (3) A one-size-fits-all approach to visual processing is suboptimal; the computational budget should adapt to sample/task demands. 

\section{Method}\label{sec:method}

\subsection{Preliminaries: Large Vision-Language Models}

Let $\mathbf{V} \in \mathbb{R}^{N_v \times d}$ and $\mathbf{T} \in \mathbb{R}^{N_t \times d}$ be the sequences of visual and text tokens, respectively, processed by an LVLM. In a standard LVLM, each transformer layer (TL) $l$ consists of a self-attention layer followed by a feed-forward network (FFN) applied to the concatenated sequence $[\mathbf{V}^{(l-1)}; \mathbf{T}^{(l-1)}]$:
\begin{equation}
    [\mathbf{V}^{(l)}; \mathbf{T}^{(l)}] = \text{TL}_l([\mathbf{V}^{(l-1)}; \mathbf{T}^{(l-1)}]).
\end{equation}
It is straightforward to observe that the self-attention operating on the concatenated sequence captures all possible image-image, image-text, and text-text interactions. Its computational cost is quadratic in the total sequence length, $O((N_v + N_t)^2 \cdot d)$. Since $N_v \ge N_t$, especially for high-resolution images, the image-image interactions dominate the inference cost.

\subsection{Vision on Request (\name)}

\begin{figure*}[!ht]
    \centering
    \includegraphics[width=\linewidth]{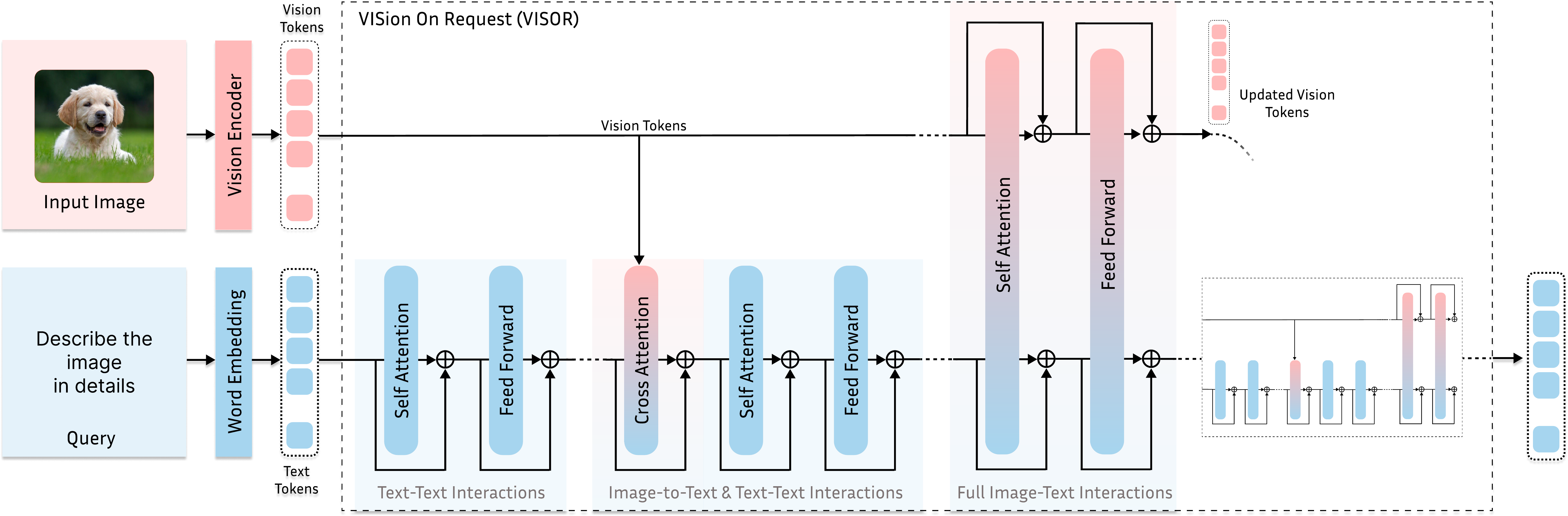}
    \caption{\textbf{Conceptual architecture of \name{}.} Visual information is sparsely injected into the language stream via a few cross-attention and self-attention layers modelling text-image and image-image interactions. Cross-attention efficiently provides visual context to the text tokens without altering the visual representations. Self-attention, while more costly, refines the visual tokens, enabling subsequent cross-attention layers to access higher-level visual features. This design strikes a balance between efficiency and representational power.}
    \label{fig:method}
    \vspace{-0.3cm}
\end{figure*}

To reduce the computational cost without performing token reduction, we propose \name{} that modifies the LVLM architecture to process visual information sparsely. The core idea is to decouple the processing of text and vision tokens. Most LLM layers operate only on text tokens. Only a few selected layers additionally integrate text-image and image-image interactions by strategically inserting a small number of cross-attention and self-attention layers, as illustrated in Fig.~\ref{fig:method}~\footnote{More precisely, self-attention models all possible interactions, including the image-image ones.}. Crucially, the inserted layers depend on sample/task complexity.

\subsubsection{Efficient Visual Context via Cross-Attention}

For many tasks, the LLM only needs to query visual features without needing to update them. Cross-attention layers provide an efficient mechanism for this, as they integrate visual information into the text processing stream without modifying the visual tokens themselves. We leverage this by having most transformer layers operate solely on text tokens. We then designate a small, uniformly distributed subset of layers, indexed by a set $\mathcal{L}_{CA}$, to perform cross-attention, allowing the text stream to efficiently query the static visual features at selected points.

Let $\mathbf{V}^{(0)}$ be the initial visual tokens from the vision encoder. For a layer $l$, the update rule is:
\begin{equation}
\begin{split}
    \mathbf{V}^{(l)} &= \mathbf{V}^{(l-1)}, \\
    \mathbf{T}^{(l)} &= 
    \begin{cases} 
        \text{TL}_l(\text{CrossAttn}(\mathbf{T}^{(l-1)}, \mathbf{V}^{(l-1)})), & \text{if } l \in \mathcal{L}_{CA} \\
        \text{TL}_l(\mathbf{T}^{(l-1)}), & \text{otherwise.}
    \end{cases}
\end{split}
\label{eq:cross-attention-only}
\end{equation}

The CrossAttn module uses text tokens as queries and visual tokens as keys and values, and its output is added residually to the text stream. Crucially, in this cross-attention-only variant, visual tokens $\mathbf{V}^{(l-1)}$ are never updated (i.e., $\mathbf{V}^{(l-1)} = \mathbf{V}^{(0)}, \forall l$), making the process highly efficient.

Finally, to ensure the vision tokens retain positional information, which is essential for spatial reasoning, inspired by \citet{chu2021conditional}, we adapt the idea of conditional positional embeddings to 1D sequences and implement them using a 1D depth-wise convolutional layer (with kernel size 7 and a padding of 3). This approach effectively captures both local and global positional information without the slower convergence issues associated with absolute or rotary positional embeddings.

\subsubsection{Refining Visual Features with Selective Self-Attention}

The cross-attention only model described in Eq.~\ref{eq:cross-attention-only} is efficient and performs well on tasks
requiring coarse visual understanding, often surpassing prior state-of-the-art methods. However, the visual tokens remain unchanged, which limits performance on tasks requiring fine-grained reasoning. To address this, we introduce a small number of full self-attention layers on the visual tokens at specific layers, indexed by a set $\mathcal{L}_{SA}$. These layers allow the model to build hierarchical visual representations.

Let us define $\mathbf{Z} = \text{CrossAttn}(\mathbf{T}^{(l-1)}, \mathbf{V}^{(l-1)})$. Then the complete update rule for a layer $l$ becomes:
\begin{equation}
\begin{split}
    &(\mathbf{V}^{(l)}, \mathbf{T}^{(l)}) =
    \begin{cases}
        \text{TL}_l([\mathbf{V}^{(l-1)}; \mathbf{T}^{(l-1)}]), & \text{if } l \in \mathcal{L}_{SA} \\
        (\mathbf{V}^{(l-1)}, \text{TL}_l(\mathbf{Z})), & \text{if } l \in \mathcal{L}_{CA} \\
        (\mathbf{V}^{(l-1)}, \text{TL}_l(\mathbf{T}^{(l-1)})), & \text{otherwise.}
    \end{cases}
\end{split}
\label{eq:cross-and-self-attention}
\end{equation}
When $l \in \mathcal{L}_{SA}$, a standard transformer layer processes both visual and text tokens, updating $\mathbf{V}^{(l-1)}$ to $\mathbf{V}^{(l)}$. Subsequent cross-attention layers ($l' > l, l' \in \mathcal{L}_{CA}$) will then use these refined visual tokens $\mathbf{V}^{(l)}$, enabling more effective context integration. In practice, we find that distributing a few cross-attention and self-attention layers uniformly across the model yields strong performance.

\subsubsection{Training a Universal Model for Adaptive Computation} \label{ssec:universal-model}
A key insight from our analysis in Sec.~\ref{sec:motivation} is that different tasks require varying amounts of visual processing. To accommodate this without training and storing multiple models, we train a single, \textit{universal} \name{} model capable of operating at various computational budgets. This is achieved by making the model robust to executing different subsets of its self-attention layers, which we refer to as \textit{configurations}. To this end, we propose the following training strategy:

\noindent\textbf{1. Bounding the configurations space.}
Given a model with $L$ total layers, we first determine the maximum number of cross-attention ($|L_{CA}|$) and self-attention ($|L_{SA}|$) layers needed to match the performance of the original dense model. Empirically, we find that setting $|L_{CA}| = |L_{SA}| = L/3$ provides a strong upper bound (see Sec.~\ref{sec:ablations}, Table~\ref{tab:config_category_comparison}). We then pre-train a \name{} model with this maximal configuration to establish a reference network.

\noindent\textbf{2. Identifying viable sub-networks.}
As the space of possible sub-networks is vast, with many configurations leading to catastrophic performance degradation due to skipping critical layers needed for certain tasks, we propose to systematically evaluate subsets from the pre-trained model to identify a set of \textit{viable configurations} - those that maintain high accuracy at least in certain cases. Moreover, as the cross-attention layers are computationally inexpensive\footnote{The FLOPs for a full self-attention layer are approx. $O((N_t + N_v)^2 d)$, whereas for a cross-attention layer, they are only $O(N_t N_v d)$.} and provide essential visual context, we opt to always execute them, only varying the number and location of self-attention layers to create different computational budgets.
Hence, we evaluate the model's performance by systematically varying the number of self-attention layers from 0 to $|L_{SA}|$, testing various subsets of the $|L_{SA}|$ layers. See Sec.~\ref{sec:ablations} for ablation results and supplementary material for more details and visualizations.

\noindent\textbf{3. Universal fine-tuning.}
Finally, inspired by~\cite{cai2019once}, we finetune the model by randomly selecting at each optimization step one of these viable configurations. This results in a universal model that works robustly for any of the configurations used during training, and hence, across a wide range of computational budgets.

\subsection{Adaptive Inference} ~\label{ssec:adaptive-inference}
As highlighted in Sec.~\ref{sec:motivation}, the amount of visual processing required varies significantly depending on the task and even across individual samples within the same benchmark. This observation indicates that a single, fixed configuration may not be optimal for all scenarios. To address this, we utilize our universal model of Sec.~\ref{ssec:universal-model} (designed to operate across a range of pre-defined computational budgets) and introduce a lightweight policy network that dynamically decides how many self-attention layers to execute for each input, enabling per-sample adaptation.

We implement this with an internal routing mechanism. A special \textit{routing} token is appended after the question, and we place an MLP layer at the block prior to the first self-attention block that is a candidate for being skipped. That MLP processes the routing token and predicts the optimal configuration for the subsequent self-attention layers.
If multiple questions are present, the model conservatively selects the configuration with the highest computational cost among the individual predictions to ensure sufficient processing capacity.

Since training a routing mechanism can be unstable~\cite{dai2022stablemoe}, we adopt an offline pseudo-labeling approach. First, we run our universal model on a training subset, logging the correctness and token-level losses for each potential layer configuration.
We then generate a pseudo-label for the subset by identifying the most efficient configuration. To do this, we first filter for configurations that achieve at least 99\% of the full model's accuracy. From this group, we select the one with the fewest layers and the lowest aggregate loss.
This chosen configuration becomes the target label for training the policy network using a standard cross-entropy loss.

\subsection{Combining Vision-on-Request with Token Reduction}
\label{ssec:combining-pia-token-reduction}

Our approach is orthogonal to existing token reduction methods and can be combined with them for further efficiency gains. To this end, we explore two strategies: (i) combining \name{} with top-performing token pruning methods~\citep{yang2025visionzip,zhang2025beyond}, and (ii) designing a simple token packing strategy that works with arbitrary token compression ratios. The latter method is coined \name-TR. Additional details can be found in the supplementary material.

\section{Experiments}~\label{sec:experiments}

We compare our method against state-of-the-art approaches on a wide range of vision-language benchmarks, covering tasks that require both coarse and fine-grained visual understanding. We show that prior methods are competitive on easy tasks, but struggle on harder tasks that require detailed visual reasoning. In contrast, our method consistently outperforms prior works across all benchmarks, particularly excelling on the challenging tasks.

\subsection{Experimental setup}~\label{ssec:experiments-setup}

\textbf{Model architecture and training details:} We build upon the open-sourced LLaVA-OV model~\citep{li2024llavaone}, which uses a SigLIP-400M~\citep{zhai2023sigmoid} vision encoder, a Qwen2~\citep{yang2024qwen2} LLM, and a 2-layer MLP connector. The vision encoder operates on $384\times384$ image patches, each patch resulting in 729 visual tokens. 
We insert cross-attention and self-attention layers uniformly across the LLM, at a maximum of 1/3 of the total layers each.
We train our model on the same datasets as LLaVA-OV, i.e., (1) the 4M pretraining knowledge data formed 
by combining synthetically labeled parts of CC3M~\citep{sharma2018conceptual}, COCO118K~\citep{lin2014microsoft}, BLIP558K~\citep{liu2024improved}, SynthDog~\citep{kim2022donut} and Evol-Instruct~\citep{chen2024allava} and (2) the LLaVA-OV Single-Image 3.2M dataset~\citep{li2024llavaone}, a high-quality mixture formed by combining over 80 datasets. We note that some of the partitions were not made available, hence, in practice, we train on a smaller set (as defined in the LLaVA-OV GitHub repository).

Our training follows a similar two-stage procedure. First, we finetune the new attention layers on the 4M knowledge dataset while freezing the rest of the model. Then, we finetune the entire model on the 3.2M high-quality dataset. Training spans 3 epochs across these stages, using the AdamW~\citep{loshchilov2017decoupled} optimizer with no weight decay, a batch size of 128, and learning rates of $1e-4$ and $1e-5$ for the first and second stages, respectively. We train on 16 MI300X GPUs using PyTorch~\citep{paszke2019pytorch} and DeepSpeed~\citep{rasley2020deepspeed}. This process applies to both universal and independent variants, except that the universal model also samples configurations during training (Sec.~\ref{ssec:universal-model}, and supplementary material for details). 

\textbf{Vision-language benchmarks:} We evaluate our models on a comprehensive set of vision-language benchmarks designed to assess diverse aspects of visual understanding. Specifically, we include the following datasets: RealWorldQA~\citep{grok15v}, ScienceQA~\citep{lu2022learn}, GQA~\citep{hudson2019gqa}, MME~\citep{zhang2021mme}, MMSTAR~\citep{chen2024we}, MMBench~\citep{liu2024mmbench}, POPE~\citep{li2023evaluating}, AI2D~\citep{kembhavi2016diagram}, ChartQA~\citep{masry2022chartqa}, TextVQA~\citep{singh2019towards}, InfoVQA~\citep{mathew2022infographicvqa}, OCRBench~\citep{liu2024ocrbench}, and DocVQA~\citep{mathew2021docvqa}. 
To better analyze the model's performance, based on the observations from Sec.~\ref{sec:motivation}, we categorize these datasets into two: \textit{easy}, which involve limited text-to-image interactions, and \textit{hard} tasks, requiring extensive text-to-image and image-to-image interactions. 

\textbf{State-of-the-art baselines:} We compare our method against several training-free and training-aware approaches. Since many existing token reduction methods are designed for different architectures (e.g., LLaVA-1.5~\citep{liu2024improved}) and focus on lower-resolution tasks, we re-implement and evaluate them under a unified setting using the same LLaVA-OV architecture and, where applicable, training data. Details of the re-implementations are provided in the supplemenatary material.

\subsection{Comparison with the state-of-the-art}

We compare our method against state-of-the-art approaches using a shared LLaVA-OV (0.5B) backbone. In addition to the numerical accuracy, we also report average FLOP savings relative to the baseline LLaVA-OV model. Note that we do not take into consideration the vision encoder FLOPs as they are common to all methods. For our method, we always use a single universal \name{} model with routing that ensures adaptability across different tasks. Table~\ref{tab:comparison_with_sota_05B} summarizes the results.

\name~achieves significant improvements in both accuracy and computational efficiency. On tasks requiring coarse visual context, \name{} matches or exceeds the performance of prior methods while achieving up to $8.6\times$ FLOP savings. For tasks demanding fine-grained visual reasoning, \name{} outperforms all baselines, including token reduction methods like VisionZip~\citep{yang2025visionzip}, HiRED~\citep{arif2025hired}, and M$^3$\citep{cai2025matryoshka}, which struggle with information bottlenecks.
When combined with token reduction techniques (\name-TR) in Sec.~\ref{ssec:combining-pia-token-reduction}, our method achieves even greater efficiency (up to $18\times$ FLOP savings), while maintaining state-of-the-art accuracy. 

See the supplementary material for more results, including comparisons using larger backbones (LLaVA-OV 1.5B).

\begin{table*}[!ht]
\centering
\caption{Comparison with state-of-the-art methods on various vision-language benchmarks. The metric used is accuracy for all datasets, except for MME where we report a score (higher is better; MME values are divided by 20 for normalization purposes).}
\label{tab:comparison_with_sota_05B}
\resizebox{\linewidth}{!}{%
\begin{tabular}{l|ccccccccc|ccccc|c}
\toprule
\multirow{7}{*}{\textbf{Method}} & \multicolumn{9}{c|}{\textit{Easy}} & \multicolumn{5}{c|}{\textit{Hard}} & \multirow{7}{*}{\textbf{\begin{tabular}[c]{@{}c@{}}Avg. \\ FLOPs\\ Savings\end{tabular}}} \\
\cmidrule(lr){2-10} \cmidrule(lr){11-16}
& \rotatebox{90}{RWorldQA} & \rotatebox{90}{SQA} & \rotatebox{90}{GQA} & \rotatebox{90}{MME} & \rotatebox{90}{MSTAR} & \rotatebox{90}{POPE} & \rotatebox{90}{TextVQA} & \rotatebox{90}{AI2D} & \rotatebox{90}{\textbf{Avg. (Easy)}} & \rotatebox{90}{ChartQA} & \rotatebox{90}{OCRBench} & \rotatebox{90}{InfoVQA} & \rotatebox{90}{DocVQA} & \rotatebox{90}{\textbf{Avg. (Hard)}}\\
\midrule
LLaVA-OV \citep{li2024llavaone} & 54.0 & 67.2 & 58.3  & 60.6 & 40.6 & 88.4 & 66.0 & 56.7 & 61.5 & 60.9 & 58.8 & 40.0 & 68.7 & 57.1 & $1.0\times$ \\
Downsample \citep{liao2025we} & 52.8
 & 67.4 & 57.4  & 62.4 & 40.0 & 86.1 & 58.4 & 56.7 & 60.2 & 49.3 & 47.9 & 27.3 & 49.5 & 43.5 & $4.0\times$ \\
VisionZip \citep{yang2025visionzip} & 51.9 & 67.0 & 53.0 & 62.1 & 38.9 & 85.5 & 46.5 & 53.2 & 57.3 & 44.0 & 27.0 & 23.9 & 36.7 & 32.9 & $5.7\times$ \\
VisionZip$^{\dagger}$ \citep{yang2025visionzip} & 54.7 & 65.8 & 55.7  & 61.9 & 39.3 & 86.9 & 55.7 & 54.8 & 59.3 & 51.2 & 45.0 & 27.2 & 48.8 & 43.1 & $5.7\times$ \\
VisPruner \citep{zhang2025beyond} & 53.3 & 65.7 & 53.2 & 60.1 & 38.1 & 85.9 & 47.8 & 53.7 & 57.2 & 44.0 & 28.3 & 25.4 & 42.7 & 35.1 & $5.7\times$ \\
SparseVLM \citep{zhang2024sparsevlm} & 51.0 & 66.7 & 48.7 & 57.2 & 38.5 & 77.6 & 62.7 & 53.9 & 57.0 & 57.2 & 43.9 & 33.1 & 62.3 & 49.1 & $4.5\times$ \\
PyramidDrop \citep{xing2025pyramiddrop} & 53.1 & 66.7 & 53.5 & 59.4 & 40.1 & 86.0 & 45.5 & 54.3 & 57.3 & 51.1 & 42.2 & 30.4 & 48.0 & 42.9 & $4.2\times$ \\
M$^3$ \citep{cai2025matryoshka} & 54.0 & 75.1 & 59.7 & 63.8  & 40.9 & 88.6 & 67.0 & 62.5 & 64.0 & 64.7 & 58.0 & 38.3 & 65.4 & 56.6 & $8.0\times$ \\
HiRED \citep{arif2025hired} & 52.6 & 66.3 & 55.4 & 61.9 & 39.2 & 86.6 & 56.8 & 55.3 & 59.3 & 47.5 & 33.9 & 26.1 & 48.4 & 39.0 & $5.0\times$ \\
\midrule
\rowcolor{lightblue}
\rowcolor{lightblue}
\name\ \textbf{(Ours)} & 54.6 & 75.3 & 61.8 & 60.5 & 40.1 & 87.6 & 67.8 & 61.5 & 63.6 & 65.2 & 61.8 & 37.6 & 68.9 & 58.4 & $8.6\times$ \\
\rowcolor{lightblue}
\name-TR \textbf{(Ours)} & 55.4 & 75.4 & 60.7 & 59.5 & 38.5 & 87.7 & 67.4 & 61.9 & 63.3 & 65.3 & 60.7 & 37.4 & 67.7 & 57.8 & $18\times$ \\
\arrayrulecolor{black}\bottomrule
\end{tabular}%
}
 \vspace{-0.3cm}
\end{table*}

\section{Ablation studies and analysis}\label{sec:ablations}

Unless otherwise specified, all ablations are performed using the LLaVA-OV (0.5B) backbone, trained on the same datasets as the main experiments. The results reported are aggregated across the two task categories for brevity.

\begin{table}[ht]
    \centering
    \caption{Accuracy comparison when combining \name{} with token reduction methods.}
    \label{tab:combining_with_token_reduction}
    \begin{tabular}{c|c|c|c}
        \toprule
        \textbf{Method} & \textbf{FLOPs sav.}  & \textbf{Easy} & \textbf{Hard} \\
        \midrule
        \name & $8.9\times$ &    63.6 & 58.4 \\
        \name-TR [2x] &  $17.8\times$   & 63.3 & 57.8 \\
        \midrule
        \multicolumn{4}{c}{approx. 4x token reduction} \\
        \midrule
        \name-TR [4x]  &  $35.0\times$   & 63.1 & 56.2 \\
        \name + VisionZip & $37.0\times$   & 63.3 & 55.3 \\
        \name + VisPruner & $39.0\times$   & 63.5 & 55.9 \\
        \bottomrule
    \end{tabular}
\end{table}

\begin{table}[ht]
    \centering
    \caption{Acc. comparison across configurations and categories.}
    \label{tab:config_category_comparison}
    \begin{tabular}{c|c|c|c || c|c|c|c}
        \toprule
        \textbf{SA} & \textbf{CA} & \textbf{Easy} & \textbf{Hard} & \textbf{SA} & \textbf{CA} & \textbf{Easy} & \textbf{Hard} \\
        \midrule
        0 & 6  & 63.3 & 51.8 & 2 & 8 & 63.2 & 55.1 \\
        0 & 8  & 63.4 & 52.0 & 4 & 8 & 63.3 & 56.9 \\
        0 & 10 & 63.5 & 52.1 & 7 & 8 & 63.9 & 58.3 \\
        \bottomrule
    \end{tabular}
    \vspace{-0.1cm}
\end{table}

\noindent\textbf{Effect of cross-attention and self-attention layers:} Herein, we analyze the impact of varying the number of cross-attention (CA) and self-attention (SA) layers on accuracy. To avoid a potential sampling bias, each configuration corresponds to an independently trained model. From Table~\ref{tab:config_category_comparison}, we can observe that: (1) cross-attention alone suffices for tasks requiring coarse visual context, with performance saturating around 8 layers; (2) Cross-attention alone is insufficient for tasks demanding fine-grained reasoning, significantly lagging behind the full model; (3) adding self-attention layers substantially boosts performance on fine-grained tasks, with a 7 layer configuration nearly matching the full model. This underscores the need for visual feature refinement in complex tasks.

\noindent\textbf{Combining \name~with token reduction methods:} Our method is orthogonal to token reduction techniques and can complement them for greater efficiency. We evaluate the impact of combining \name{} with token reduction methods like VisionZip~\citep{yang2025visionzip}, VisPruner~\citep{zhang2025beyond}, and our token packing strategy (\name-TR) from Sec.~\ref{ssec:combining-pia-token-reduction}, under varying reduction rates. As Table~\ref{tab:combining_with_token_reduction} shows, our approach can benefit from token reduction, achieving up to $35\times$ FLOPs savings with only a minor drop in accuracy. Notably, more aggressive token reduction rates (e.g., $4\times>$) lead to larger performance drops on hard tasks, as the information bottleneck becomes more pronounced.

\noindent\textbf{Independent vs universal model training:} Our final model is trained to support multiple configurations, enabling dynamic adjustment of computational cost during inference. Herein, we analyze the performance trade-offs of this universal training approach compared to independently training models at a few selected budgets. As Table~\ref{tab:independent_vs_universal} indicates, the universal model matches and, surprisingly, surpasses the performance of independently trained models across different budgets, while providing the flexibility of adaptive inference. This suggests that the universal training approach also acts as a form of regularization, improving generalization across configurations.

\begin{table}
\centering
\vspace{-0.2cm}
\caption{Accuracy comparison between independently trained models and a universally trained model supporting multiple configurations. Both model variants use the same fixed configuration for all samples.}
\label{tab:independent_vs_universal}
\resizebox{0.75\linewidth}{!}{%
\begin{tabular}{c|c|c|c|c}
\toprule
\textbf{SA} & \textbf{CA} & \textbf{Universal}   & \textbf{Easy} & \textbf{Hard}  \\
\midrule
\multirow{2}{*}{2} & \multirow{2}{*}{8} & $\times$ & 63.1  & 55.1 \\
& & \checkmark & 63.8 & 56.3 \\
\midrule
\multirow{2}{*}{7} & \multirow{2}{*}{8} & $\times$ & 63.6 & 58.3  \\  
& & \checkmark & 64.2 & 59.1  \\
\bottomrule
\end{tabular}%
}
\vspace{-0.3cm}
\end{table}

\noindent\textbf{Efficiency analysis:} We analyze the computational efficiency of \name{} by measuring the number of floating-point operations (FLOPs) required for inference. The primary source of savings in our method comes from replacing the expensive full self-attention over all tokens with either text-only self-attention or a cheaper cross-attention mechanism in most layers. The cost of a standard transformer layer is quadratic in the total sequence length, $O((N_t + N_v)^2 d + (N_t+N_v)d^2)$, which is dominated by the large number of visual tokens $N_v$. In contrast, our cross-attention layers have a cost of only $O(N_t N_v d)$, and text-only layers are independent of $N_v$. The full, expensive self-attention is computed only in a small, selective subset of layers.

Fig.~\ref{fig:efficiency-plot} illustrates the computational cost - accuracy tradeoff. Our approach significantly reduces the FLOPs compared to the baseline LLaVA-OV model while offering a better accuracy-efficiency trade-off compared to all prior methods. Note that the FLOPs measured here only account for the transformer layers, excluding the vision encoder, which is common to all methods.

We also measure actual inference speedups on real hardware (MI300X GPUs). As our approach is compatible to existing efficient attention implementations (e.g., flash attention~\cite{dao2022flashattention}), we observe a good correlation between actual speedup and FLOP savings: A full LLaVA-OV model takes $0.0738$sec/sample, a token pruning solution (i.e., VisionZip, VisPruner) at 8x reduction factor - $0.0274$ sec/sample, while \name~takes $0.0384$ sec/sample at a 8CA-7SA configuration and  $0.0261$ sec/sample at a 8CA-2SA. The numbers are reported at the max batch size that allows all methods to fit in memory. See supplementary material for additional analysis.
Finally, in terms of model size,  the CA layers from \name{} introduces a small overhead (less than 7.5\%) compared to the baseline LLaVA-OV model, concentrated in the linear projection layers.

\section{Conclusion}

In this work, we proposed Vision-on-Request (\name), a method that reduces the inference cost in LVLMs by sparsifying the image-text and image-image interactions without discarding visual information (as in previous token reduction methods). \name{} uses efficient cross-attention to model text-image interactions and few selective self-attention layers for visual feature refinement necessary for fine-grained visual understanding and reasoning. 
\name{} trains a single universal network on a range of computational budgets and then, during inference, utilizes lightweight policies that dynamically allocate visual computation based on per-task or per-sample complexity. We show that \name{} drastically improves efficiency, and outperforms state-of-the-art token compression methods across a wide range of benchmarks, especially for challenging tasks that require detailed visual understanding.

{
    \small
    \bibliographystyle{ieeenat_fullname}
    \bibliography{main}
}

\appendix

  \section{Identifying promising configurations for adaptive training and
    inference}

    Considering a model with $L_{SA}$ self-attention layers, we can create
    $2^{L_{SA}}$ different configurations by choosing to execute or skip each
    self-attention layer. This results in a vast configuration space, making it
    somewhat impractical to evaluate all of them. Moreover, many configurations
    may lead to catastrophic performance degradation, as they may skip critical layers
    needed for certain tasks. Hence, to facilitate the training process, we seek
    to identify a subset of promising configurations that maintain high
    performance. This subset can then be used for adaptive training and inference.

\begin{figure*}[!ht]
        \centering
        \includegraphics[width=1.0\linewidth]{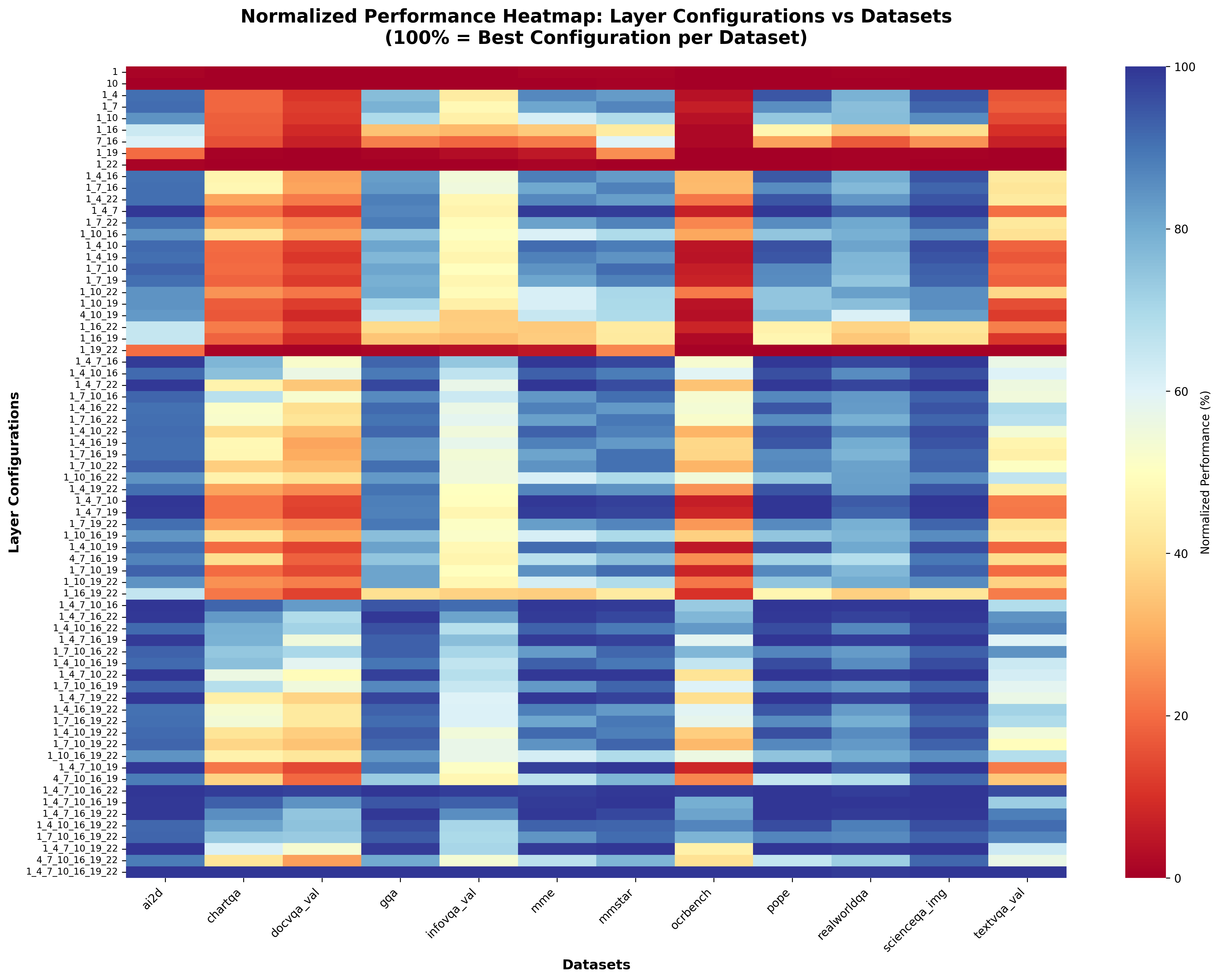}
        \caption{Performance heatmap for different configurations across
        datasets. Each row represents a configuration, and each column
        corresponds to a dataset. The color intensity indicates the relative accuracy
        achieved by that configuration on the respective dataset.}
        \label{fig:performance_heatmap}
    \end{figure*}

    Figure~\ref{fig:performance_heatmap} visualizes the performance of a representative
    subset of configurations on different datasets, with each row representing a
    configuration and each column a dataset. The color intensity indicates the relative
    accuracy achieved by that configuration on the respective dataset. From this
    visualization, we can identify that: (1) dropping the 1st layer leads to
    significant performance degradation across all datasets, indicating its critical
    role; (2) configurations with very few self-attention layers (e.g., 1 or 2)
    perform poorly on complex tasks, while those with more layers generally yield
    better results; (3) less vision intensive tasks generally prefer a
    configuration close to early-exit while more complex tasks benefit from a
    uniform distribution of self-attention layers.

    Based on these observations, for a 0.5B model, we subsequently select the following
    configurations, where each number denotes the location at which a self-attention
    layer is executed for the vision tokens:\texttt{[1,4], [1,7], [1,4,7], [1,4,16],
    [1,7,16], [1,10,16], [1,4,7,16], [1,4,7,22], [1,4,10,16], [1,4,7,10,16], [1,4,7,16,22],
    [1,4,7,10,16,22], [1,4,7,10,16,19,22]}.

    \section{Per-dataset saving rate}

    In the main manuscript, for brevity, we report the computational savings
    aggregated across all datasets. Herein, we provide a more detailed per-dataset
    analysis. Table~\ref{tab:per_dataset_comparison_with_indiv_flops} summarizes
    the results. For each variant, the top row indicates the accuracy, while the
    bottom row shows the FLOPs savings relative to the baseline LLaVA-OV model. The
    same general pattern holds: easy tasks can be solved with very few self-attention
    layers, while hard tasks require more layers for optimal performance. Our
    router is able to correctly identify this trend.

    \begin{table*}
        [!htbp]
        \centering
        \caption{Per-dataset saving rates. We compare our method against state-of-the-art
        approaches using a shared LLaVA-OV (0.5B) backbone. For each method, the
        top row indicates the accuracy, while the bottom row shows the FLOPs savings
        relative to the baseline LLaVA-OV model. The metrics used are accuracy
        for most datasets, except for MME where we report a score (higher is
        better).}
        \label{tab:per_dataset_comparison_with_indiv_flops}
        \resizebox{\linewidth}{!}{%
        \begin{tabular}{l|cccccccccccc}
            \toprule \textbf{Method}           & \textbf{RWQA} & \textbf{SQA} & \textbf{GQA} & \textbf{MME} & \textbf{MMSTAR} & \textbf{POPE} & \textbf{TxtVQA} & \textbf{AI2D} & \textbf{CQA} & \textbf{OCRB} & \textbf{InfoVQA} & \textbf{DocVQA} \\
            \midrule \multirow{2}{*}{\name}    & 54.6          & 75.3         & 61.8         & 60.5         & 40.1            & 87.6          & 67.8            & 61.5          & 65.2         & 61.8          & 37.6             & 68.9            \\
                                               & $8.4\times$   & $8\times$    & $12\times$   & $8.3\times$  & $10\times$      & $12\times$    & $6.3\times$     & $12\times$    & $8\times$    & $6.6\times$   & $6\times$        & $6\times$       \\
            \midrule \multirow{2}{*}{\name-TR} & 55.4          & 75.4         & 60.7         & 59.5         & 38.5            & 87.7          & 67.4            & 61.9          & 65.3         & 60.7          & 37.4             & 67.7            \\
                                               & $24\times$    & $16\times$   & $24\times$   & $17.7\times$ & $17.7\times$    & $24\times$    & $14.5\times$    & $24\times$    & $16\times$   & $15\times$    & $12\times$       & $12\times$      \\
            \arrayrulecolor{black}\bottomrule
        \end{tabular}%
        }
    \end{table*}

\begin{figure*}[!ht]
        \centering
        \includegraphics[width=1.0\linewidth]{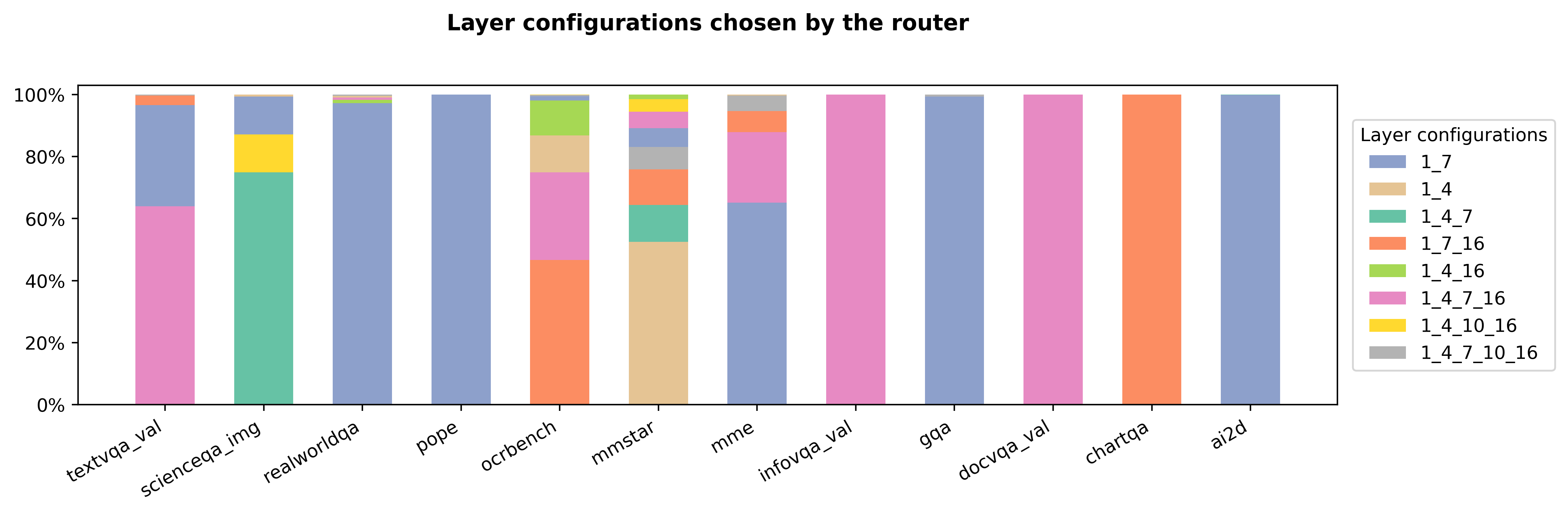}
        \caption{Layer configuration assignments made by the router for each
        test dataset.}
        \label{fig:router_choices}
    \end{figure*}

    To provide further insight into the routing mechanism's behaviour, we
    present in Figure~\ref{fig:router_choices} the layer configurations it
    chooses for each test set dataset. Two significant observations can be made from
    this figure: a) the routing mechanism is largely consistent with regard to the
    computational budget allocated for each dataset, as the configurations
    chosen for each dataset tend to have a similar number of layers, and b)
    despite the fact that the original labels are defined in a per-dataset basis,
    the routing mechanism indeed operates on a per-sample basis, which makes it
    adaptive to individual samples' complexity.

    \section{Performance across all individual configurations}

\begin{figure*}[!ht]
        \centering
        \includegraphics[width=1.0\linewidth]{
            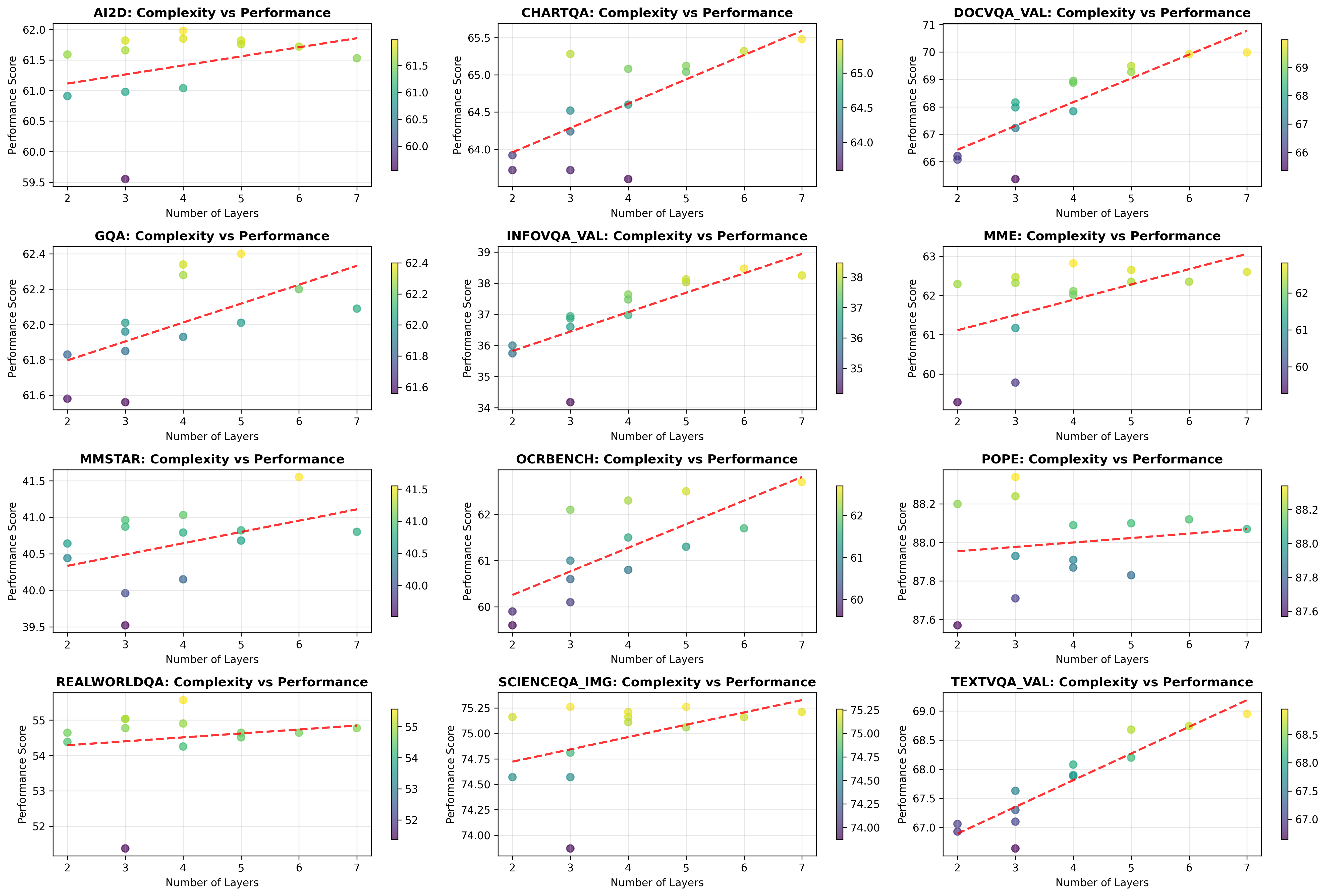
        }
        \caption{Performance for various~\name{} configurations.}
        \label{fig:perf_for_all_configs}
    \end{figure*}

    Our universal model is trained by randomly sampling a viable configuration during
    training. To evaluate the effectiveness of each configuration, Figure~\ref{fig:perf_for_all_configs}
    illustrates their performance across all downstream tasks. On the easy
    partition, most configurations achieve similar performance regardless of their
    computational cost. In contrast, for challenging tasks, performance improves
    almost linearly with the computational budget, highlighting once more the importance
    of additional self-attention layers for fine-grained reasoning.

    \section{Oracle performance analysis}

\begin{figure*}[!ht]
        \centering
        \includegraphics[width=1.0\linewidth]{
            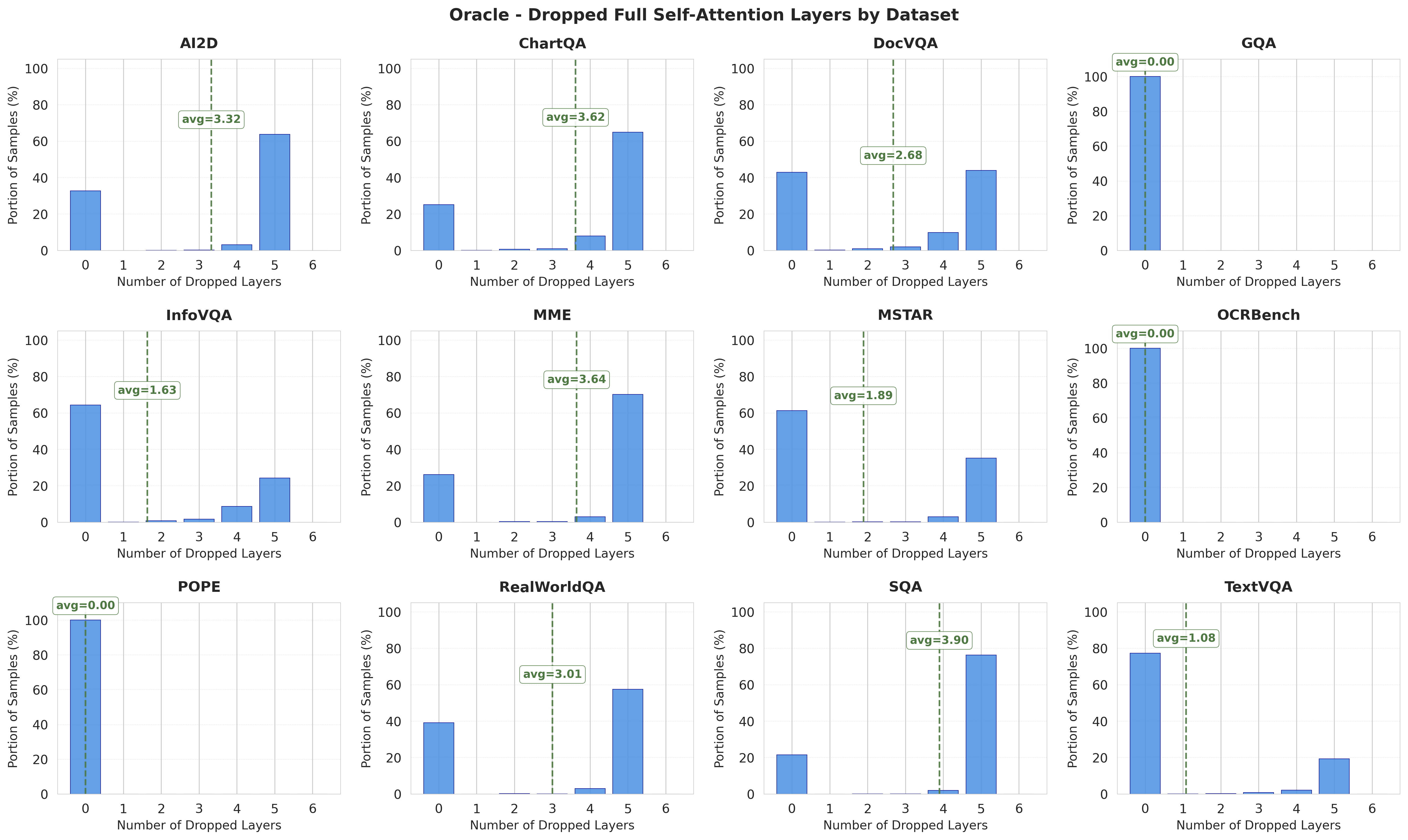
        }
        \caption{\name{} oracle: smallest amount of layers that maximizes
        accuracy.}
        \label{fig:lowest_flops_oracle}
    \end{figure*}

    To assess the potential of our adaptive inference mechanism, we conduct an
    oracle analysis where we select the optimal configuration for each sample
    from our predefined set. Figure~\ref{fig:lowest_flops_oracle} illustrates the
    distribution of the selected configurations across all samples such that the
    overall accuracy is maximized. These results reinforce the conclusion that
    most samples can be accurately processed using configurations with very few
    self-attention layers, while hard tasks require more layers for optimal performance.

    To find the optimal configuration per sample, we compare the per-config
    generations token by token to the tokenized ground-truth, and score it as the
    number of matches up to the first incorrect match. Then, we select the configuration
    with the most matches. If multiple ones have the same score, we select the one
    with the minimum number of SA layers (i.e., the one with the most dropped
    layers).

    \section{Additional discussion on efficiency}

    \begin{figure}
        \centering
        \includegraphics[width=0.44\textwidth]{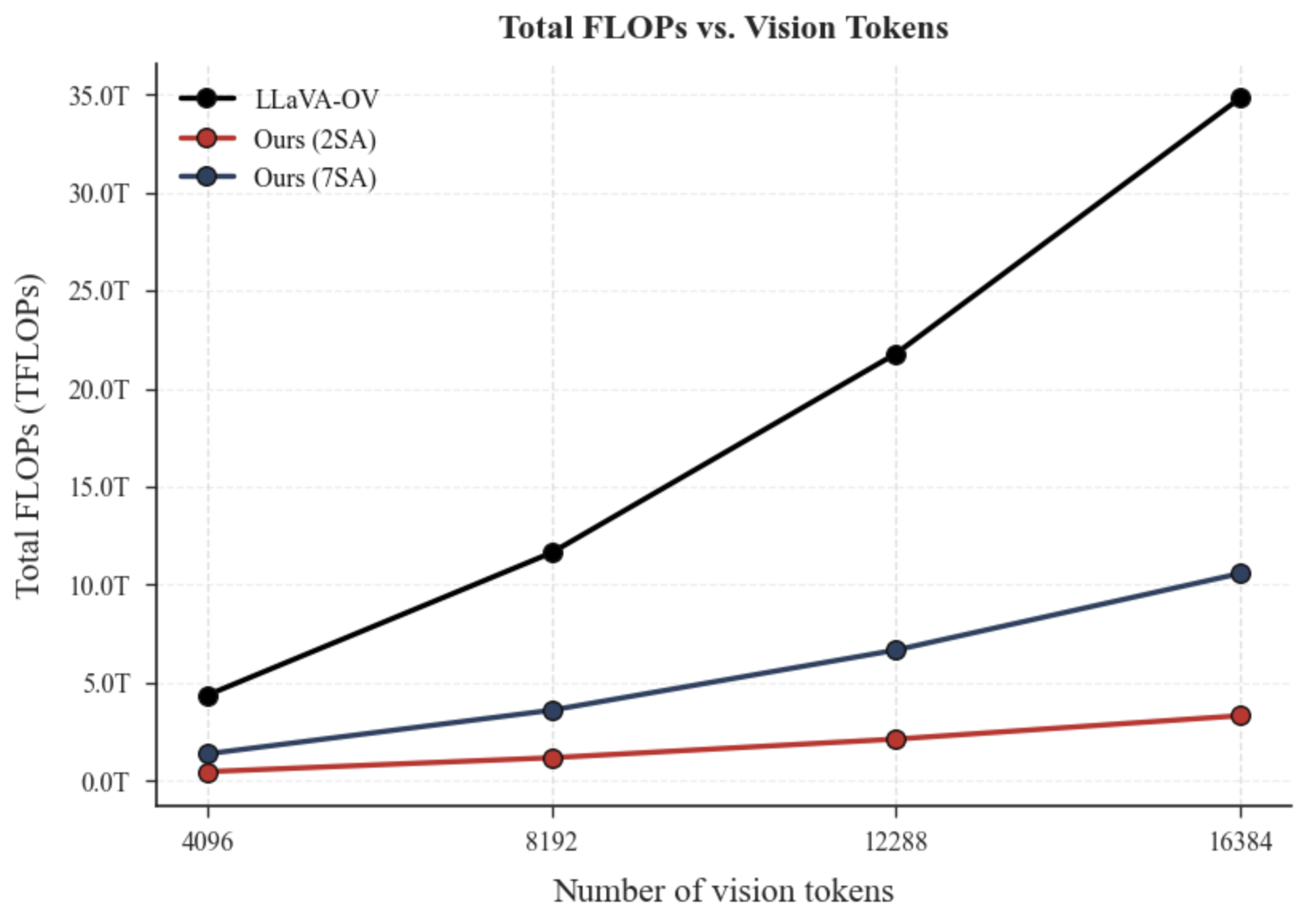}
        \caption{\textbf{Efficiency comparison} - number of FLOPS vs. vision
        sequence length.}
        \label{fig:efficiency-plot}
    \end{figure}

    In addition to the discussion from the main paper, in Fig.~\ref{fig:efficiency-plot}
    we illustrate the computational cost as a function of the visual sequence
    length. Our approach significantly reduces the FLOPs compared to the baseline
    LLaVA-OV model, with longer sequences benefiting more from the reduction. Note
    that the FLOPs measured here only account for the transformer layers,
    excluding the vision encoder which is common to all methods.

    \section{Reduced number of tokens vs reduced attention}

    In the main paper, we demonstrated that intermittent attention (\name) is
    effective for hard tasks and orthogonal to token reduction methods. Here, we
    conduct a head-to-head comparison of these two paradigms under a similar
    FLOPs reduction rate of approximately 16$\times$. We use our \name-TR
    variant and compare it against two token reduction methods, M$^{3}$ and
    VisPruner. To ensure a fair comparison, all models were finetuned end-to-end
    using the same training procedure, with the only difference being how the amount
    of input vision tokens is reduced. As shown in Table~\ref{tab:reduced_tokens_vs_reduced_attention},
    while all methods perform well on easy datasets, our approach maintains a
    significant performance advantage on harder tasks, highlighting the benefits
    of preserving a larger visual context over aggressive token reduction.

    \begin{table*}
        [!ht]
        \centering
        \caption{Comparison on various vision-language benchmarks for Qwen2-VL-2B
        LVLM.}
        \label{tab:qwenvl_comparison_full}
        \resizebox{\linewidth}{!}{%
        \begin{tabular}{l|ccccccccc|ccccc}
            \toprule \multirow{2}{*}{\textbf{Method}} & \multicolumn{9}{c|}{\textit{Easy}} & \multicolumn{5}{c}{\textit{Hard}} \\
            \cmidrule(lr){2-10} \cmidrule(lr){11-15}  & \rotatebox{90}{RealWorldQA}        & \rotatebox{90}{SQA}              & \rotatebox{90}{GQA} & \rotatebox{90}{MME} & \rotatebox{90}{MSTAR} & \rotatebox{90}{POPE} & \rotatebox{90}{TextVQA} & \rotatebox{90}{AI2D} & \rotatebox{90}{\textbf{Avg. (Easy)}} & \rotatebox{90}{ChartQA} & \rotatebox{90}{OCRBench} & \rotatebox{90}{InfoVQA} & \rotatebox{90}{DocVQA} & \rotatebox{90}{\textbf{Avg. (Hard)}} \\
            \midrule    
            Qwen2-VL-2B                               & 61.1                               & 78.1                             & 60.2                & 75.3                & 43.6                  & 87.7                 & 79.3                    & 70.4                 & 69.5                                 & 75.0                    & 77.5                     & 64.1                    & 89.5                   & 76.5                                 \\
            \name                                     & 60.0                               & 82.1                             & 60.8                & 75.1                & 49.2                  & 88.9                 & 76.1                    & 75.0                 & 70.9                                 & 78.1                    & 76.0                     & 55.9                    & 87.3                   & 74.3                                 \\
            \bottomrule
        \end{tabular}%
        }
    \end{table*}

    \begin{table*}
        [!ht]
        \centering
        \caption{Comparison between token reduction vs intermittent attention (\name).}
        \label{tab:reduced_tokens_vs_reduced_attention}
        \resizebox{\linewidth}{!}{%
        \begin{tabular}{l|ccccccccc|ccccc|c}
            \toprule \multirow{7}{*}{\textbf{Method}} & \multicolumn{9}{c|}{\textit{Easy}} & \multicolumn{5}{c|}{\textit{Hard}} & \multirow{7}{*}{\textbf{\begin{tabular}[c]{@{}c@{}}Avg. \\ FLOPs\\ Savings\end{tabular}}} \\
            \cmidrule(lr){2-10} \cmidrule(lr){11-15}  & \rotatebox{90}{RealWorldQA}        & \rotatebox{90}{SQA}                & \rotatebox{90}{GQA}                                                                      & \rotatebox{90}{MME} & \rotatebox{90}{MSTAR} & \rotatebox{90}{POPE} & \rotatebox{90}{TextVQA} & \rotatebox{90}{AI2D} & \rotatebox{90}{\textbf{Avg. (Easy)}} & \rotatebox{90}{ChartQA} & \rotatebox{90}{OCRBench} & \rotatebox{90}{InfoVQA} & \rotatebox{90}{DocVQA} & \rotatebox{90}{\textbf{Avg. (Hard)}} \\
            \midrule LLaVA-OV \citep{li2024llavaone}  & 54.0                               & 67.2                               & 58.3                                                                                     & 60.6                & 40.6                  & 88.4                 & 66.0                    & 56.7                 & 61.5                                 & 60.9                    & 58.8                     & 40.0                    & 68.7                   & 57.1                                & $1.0\times$  \\
            + $M^{3}$                                 & 53.6                               & 75.1                               & 59.1                                                                                     & 63.6                & 41.5                  & 88.1                 & 65.5                    & 62.2                 & 63.6                                 & 62.9                    & 54.0                     & 34.7                    & 58.0                   & 52.4                                & $16.0\times$ \\
            + VisPruner train.                        & 53.1                               & 69.2                               & 56.7                                                                                     & 60.9                & 36.9                  & 86.7                 & 55.4                    & 56.3                 & 59.4                                 & 52.4                    & 44.2                     & 28.2                    & 52.9                   & 44.4                                & $16.0\times$ \\
            \midrule \name-TR \textbf{(Ours)}         & 55.4                               & 75.4                               & 60.7                                                                                     & 59.5                & 38.5                  & 87.7                 & 67.4                    & 61.9                 & 63.3                                 & 65.3                    & 60.7                     & 37.4                    & 67.7                   & 57.8                                & $18\times$   \\
            \arrayrulecolor{black}\bottomrule
        \end{tabular}%
        }
    \end{table*}

    \section{Routing mechanism generalization}
    ~\label{ssec:routing-generalization}

    As described in the main manuscript, the internal routing mechanism
    responsible for deciding the optimal configuration of self-attention layers
    to be used for each sample is trained using offline, per-dataset labels extracted
    from our training set. In this subsection, in order to investigate how the router
    performs on unseen data, we train it while excluding from its train set
    three datasets (AI2D, DocVQA, and GQA), and evaluate it on vision-language benchmarks
    including those datasets. In Table~\ref{tab:routing_generalization}, we present
    the outcomes of this experiment (\name-TR-excl) and contrast them with the router
    when trained on the full train set (\name-TR). Our results demonstrate that,
    even though, naturally, the model behavior is changed, this does not lead to
    a drop of performance in general or in the excluded datasets in particular, thereby
    indicating that \name{} is robust and can handle samples outside the train
    set's distribution.

    \begin{table*}
        [t]
        \centering
        \caption{Results for the routing mechanism trained on the full train set
        (\name-TR), contrasted with training excluding samples from AI2D, DocVQA
        and GQA (\name-TR-excl).}
        \label{tab:routing_generalization} \resizebox{\linewidth}{!}{%
        \begin{tabular}{l|cccccccccccc}
            \toprule \textbf{Method}                & \textbf{RWQA} & \textbf{SQA} & \textbf{GQA} & \textbf{MME} & \textbf{MMSTAR} & \textbf{POPE} & \textbf{TxtVQA} & \textbf{AI2D} & \textbf{CQA} & \textbf{OCRB} & \textbf{InfoVQA} & \textbf{DocVQA} \\
            \midrule \multirow{2}{*}{\name-TR}      & 55.4          & 75.4         & 60.7         & 59.5         & 38.5            & 87.7          & 67.4            & 61.9          & 65.3         & 60.7          & 37.4             & 67.7            \\
                                                    & $24\times$    & $16\times$   & $24\times$   & $17.7\times$ & $17.7\times$    & $24\times$    & $14.5\times$    & $24\times$    & $16\times$   & $15\times$    & $12\times$       & $12\times$      \\
            \midrule \multirow{2}{*}{\name-TR-excl} & 55.5          & 75.6         & 61.6         & 59.2         & 38.3            & 88.3          & 66.9            & 62.0          & 65.0         & 61.1          & 37.2             & 67.6            \\
                                                    & $14.5\times$  & $15.5\times$ & $9.8\times$  & $11.4\times$ & $20\times$      & $9.8\times$   & $12\times$      & $14.5\times$  & $16\times$   & $14.1\times$  & $12\times$       & $12\times$      \\
            \arrayrulecolor{black}\bottomrule
        \end{tabular}%
        }
    \end{table*}

    \section{Additional comparison with the state-of-the-art}

    In the main manuscript, we compared our method against state-of-the-art approaches
    using a shared LLaVA-OV (0.5B) backbone. Herein, we provide additional
    results using a larger LLaVA-OV (1.5B) backbone and on a different architecture:
    QwenVL (2B).

    For the 1.5B case, as no official 1.5B variant is openly available, we re-trained
    it ourselves fully using the same procedure as described in~\citet{li2024llavaone}.
    Additionally, we've also re-implemented the baselines under the same unified
    setting. As the results from Table~\ref{tab:comparison_with_sota_15B} show, the
    same conclusions hold and our method continues to outperform existing approaches
    on the \textit{hard} partition of the dataset, while providing significant efficiency
    gains.

    For QwenVL2 (2B) we report results in Table~\ref{tab:qwenvl_comparison_full}
    - the same conclusions hold, with our approach largely matching the full
    model's performance using significantly fewer FLOPs. We note however, that
    in this case, as we don't have access to the full QwenVL training data, our
    method is at disadvantage. In practice this difference results in disproportionate
    swings for certain datasets (e.g: InfoVQA, MMSTAR).

    \begin{table*}
        [!ht]
        \centering
        \caption{Comparison with state-of-the-art models on various vision-language
        benchmarks using a LLaVA-OV 1.5B backbone. The metrics used are accuracy
        for most datasets, except for MME where we report a score (higher is
        better). MME values are divided by 20.}
        \label{tab:comparison_with_sota_15B}
        \resizebox{\linewidth}{!}{%
        \begin{tabular}{l|ccccccccc|ccccc|c}
            \toprule \multirow{7}{*}{\textbf{Method}}       & \multicolumn{9}{c|}{\textit{Easy}} & \multicolumn{5}{c|}{\textit{Hard}} & \multirow{7}{*}{\textbf{\begin{tabular}[c]{@{}c@{}}Avg. \\ FLOPs\\ Savings\end{tabular}}} \\
            \cmidrule(lr){2-10} \cmidrule(lr){11-15}        & \rotatebox{90}{RealWorldQA}        & \rotatebox{90}{SQA}                & \rotatebox{90}{GQA}                                                                      & \rotatebox{90}{MME} & \rotatebox{90}{MMSTAR} & \rotatebox{90}{POPE} & \rotatebox{90}{TextVQA} & \rotatebox{90}{AI2D} & \rotatebox{90}{\textbf{Avg. (Easy)}} & \rotatebox{90}{ChartQA} & \rotatebox{90}{OCRBench} & \rotatebox{90}{InfoVQA} & \rotatebox{90}{DocVQA} & \rotatebox{90}{\textbf{Avg. (Hard)}} &             \\
            \midrule LLaVA-OV \citep{li2024llavaone}        & 57.3                               & 75.6                               & 58.6                                                                                     & 63.7                & 40.5                   & 88.1                 & 69.9                    & 60.3                 & 64.3                                 & 64.2                    & 60.3                     & 46.7                    & 76.6                   & 62.0                                 & $1.0\times$ \\
            Downsample \citep{liao2025we}                   & 53.6                               & 76.9                               & 58.4                                                                                     & 62.5                & 39.2                   & 86.1                 & 59.5                    & 60.0                 & 62.0                                 & 46.7                    & 48.0                     & 29.1                    & 52.6                   & 44.1                                 & $4.0\times$ \\
            VisionZip \citep{yang2025visionzip}             & 52.7                               & 74.8                               & 54.2                                                                                     & 64.5                & 38.3                   & 86.2                 & 54.9                    & 57.2                 & 60.4                                 & 37.8                    & 33.8                     & 26.3                    & 44.7                   & 35.7                                 & $5.7\times$ \\
            VisionZip$^{\dagger}$ \citep{yang2025visionzip} & 54.6                               & 76.2                               & 57.2                                                                                     & 63.9                & 38.1                   & 87.0                 & 62.0                    & 58.2                 & 62.2                                 & 50.0                    & 44.3                     & 30.8                    & 55.1                   & 45.1                                 & $5.7\times$ \\
            SparseVLM \citep{zhang2024sparsevlm}            & 52.0                               & 74.8                               & 50.3                                                                                     & 58.4                & 34.5                   & 82.3                 & 60.6                    & 57.1                 & 58.7                                 & 51.3                    & 38.9                     & 34.9                    & 57.9                   & 45.7                                 & $4.5\times$ \\
            PyramidDrop \citep{xing2025pyramiddrop}         & 56.7                               & 76.0                               & 55.9                                                                                     & 63.7                & 37.5                   & 87.2                 & 60.3                    & 62.1                 & 59.1                                 & 47.9                    & 31.9                     & 34.2                    & 54.5                   & 42.1                                 & $4.6\times$ \\
            VisPruner \citep{zhang2025beyond}               & 52.9                               & 76.0                               & 53.4                                                                                     & 62.3                & 38.4                   & 83.3                 & 54.8                    & 57.5                 & 59.8                                 & 40.0                    & 34.3                     & 28.4                    & 51.2                   & 38.5                                 & $5.7\times$ \\
            HiRED \citep{arif2025hired}                     & 54.8                               & 76.0                               & 56.3                                                                                     & 63.2                & 39.3                   & 85.9                 & 59.8                    & 58.2                 & 61.7                                 & 44.1                    & 42.0                     & 27.6                    & 52.3                   & 41.5                                 & $5.0\times$ \\
            \midrule \rowcolor{lightblue}                   
            \name \textbf{(Ours)}                           & 57.5                               & 84.7                               & 62.5                                                                                     & 68.7                & 44.9                   & 88.6                 & 68.6                    & 71.7                 & 68.3                                 & 68.3                    & 63.7                     & 44.4                    & 75.3                   & 62.9                                 & $6.6\times$ \\
            \arrayrulecolor{black}\bottomrule
        \end{tabular}%
        }
    \end{table*}

    \begin{table*}
        [t]
        \centering
        \caption{\name\ LLava-OV 0.5B results when training with different portions
        of the training data. We report performance when using the full training
        set (100\%) and a reduced subset (50\%).}
        \label{tab:reduced_data} \resizebox{\linewidth}{!}{%
        \begin{tabular}{l|l|cccccccccccc}
            \toprule \textbf{Method}         & \textbf{Training data \%} & \textbf{RWQA} & \textbf{SQA} & \textbf{GQA} & \textbf{MME} & \textbf{MMSTAR} & \textbf{POPE} & \textbf{TxtVQA} & \textbf{AI2D} & \textbf{CQA} & \textbf{OCRB} & \textbf{InfoVQA} & \textbf{DocVQA} \\
            \midrule \name                   & 100\%                     & 54.6          & 75.3         & 61.8         & 60.5         & 40.1            & 87.6          & 67.8            & 61.5          & 65.2         & 61.8          & 37.6             & 68.9            \\
            \name                            & 50\%                      & 56.9          & 72.3         & 60.1         & 60.2         & 38.6            & 88.3          & 65.1            & 57.3          & 62.2         & 58.3          & 33.8             & 62.7            \\
            \arrayrulecolor{black}\bottomrule
        \end{tabular}%
        }
    \end{table*}

    \begin{table*}
        [!ht]
        \centering
        \caption{Comparison between LLaVA-OV 0.5B and \name{} with FastVLM
        vision encoder. \textit{Avg. FLOPs Savings} refers only to LLM FLOPs and
        does not include the cost of vision encoding.}
        \label{tab:fastvlm}
        \resizebox{\linewidth}{!}{%
        \begin{tabular}{l|l|ccccccccc|ccccc|c}
            \toprule \multirow{7}{*}{\textbf{Method}} & \multirow{7}{*}{\textbf{\begin{tabular}[c]{@{}c@{}}\textbf{Vision}\\ \textbf{Encoder}\end{tabular}}} & \multicolumn{9}{c|}{\textit{Easy}} & \multicolumn{5}{c|}{\textit{Hard}} & \multirow{7}{*}{\textbf{\begin{tabular}[c]{@{}c@{}}Avg. \\ FLOPs\\ Savings\end{tabular}}} \\
            \cmidrule(lr){3-11} \cmidrule(lr){12-16}  &                                                                                                      & \rotatebox{90}{RealWorldQA}        & \rotatebox{90}{SQA}                & \rotatebox{90}{GQA}                                                                      & \rotatebox{90}{MME} & \rotatebox{90}{MSTAR} & \rotatebox{90}{POPE} & \rotatebox{90}{TextVQA} & \rotatebox{90}{AI2D} & \rotatebox{90}{\textbf{Avg. (Easy)}} & \rotatebox{90}{ChartQA} & \rotatebox{90}{OCRBench} & \rotatebox{90}{InfoVQA} & \rotatebox{90}{DocVQA} & \rotatebox{90}{\textbf{Avg. (Hard)}} \\
            \midrule LLaVA-OV                         & SigLIP-400M                                                                                          & 54.0                               & 67.2                               & 58.3                                                                                     & 60.6                & 40.6                  & 88.4                 & 66.0                    & 56.7                 & 61.5                                 & 60.9                    & 58.8                     & 40.0                    & 68.7                   & 57.1                                & $1\times$  \\
            \midrule LLaVA-OV                         & FastVLM                                                                                              & 52.5                               & 67.4                               & 56.3                                                                                     & 58.4                & 35.8                  & 87.7                 & 61.3                    & 51.5                 & 58.9                                 & 57.9                    & 47.6                     & 32.5                    & 51.1                   & 47.3                                & $20\times$ \\
            \name                                     & FastVLM                                                                                              & 52.3                               & 73.7                               & 56.3                                                                                     & 58.8                & 37.7                  & 86.7                 & 61.9                    & 55.8                 & 60.4                                 & 63.0                    & 49.0                     & 33.4                    & 53.1                   & 49.6                                & $60\times$ \\
            \arrayrulecolor{black}\bottomrule
        \end{tabular}%
        }
    \end{table*}

    \section{Additional details regarding Token Packing}

    To further enhance and validate the efficacy of our approach in the main
    section, we introduce a light adaptation for token packing, capable of
    working at non-power-of-two reduction rates.

    Specifically, after the vision encoder, we reshape the patch embeddings back
    to their 2D spatial grid, interpolate the grid by a factor of $1/\sqrt{2}$ along
    each spatial dimension for a $2\times$ compression ratio, and then apply a
    space-to-depth transformation (pixel shuffle). This deterministically halves
    the number of visual tokens with minimal information loss and no added parameters,
    complementing the computational savings from our sparse attention design. Note
    that the interpolation factor can be adjusted upwards to accommodate arbitrary
    reduction rates.

    \subsection{Multi-image performance}

    To further validate the effectiveness of our approach on multi-image inputs,
    we finetune VISOR on the LLaVA-OV multi-image dataset and compare the performance
    of the resulting model with LLaVA on the MUIR, Blink, and MMIU benchmarks.
    As shown in Table~\ref{tab:multi_image}, VISOR matches or outperforms LLaVA-OV
    despite being over $3\times$ faster.

    \begin{table}[ht]
        \centering
        \caption{Multi-image performance comparison.} 
        \label{tab:multi_image} 
        \resizebox{0.75\linewidth}{!}{
        \begin{tabular}{c|ccc}
            \toprule Method & MUIR & Blink & MMIU \\
            \midrule LLaVA  & 26.8 & 40.4  & 34.2 \\
            VISOR           & 28.1 & 39.0  & 34.5 \\
            \bottomrule
        \end{tabular}
        }
    \end{table}

    \section{Scaling \name{} with Training Data Size}

    To showcase the ability of our approach to scale with increased data, we
    train \name{} using different portions of the original training set. In
    Table~\ref{tab:reduced_data}, we compare the performance of our method when trained
    on the full dataset (100\%) versus a randomly sampled subset containing half
    of the data (50\%). As observed, reducing the amount of training data generally
    causes a decline in performance. Notably, this degradation is primarily concentrated
    on harder, vision-intensive tasks (e.g., DocVQA, InfoVQA). In contrast, the performance
    on simpler datasets (such as MME, POPE, and RWQA) remains similar. These
    results highlight that \name{} effectively leverages larger training corpuses
    to progressively improve its accuracy.

    \section{Using \name{} with FastVLM}
    \citet{vasu2025fastvlm} introduce FastVLM, an efficient LVLM that uses the FastVITHD~\cite{vasu2023fastvit}
    vision backbone to encode high-resolution inputs efficiently. The FastVLM
    vision encoder uses a hybrid convolution–attention architecture and outputs
    a reduced number of visual tokens. While the standard LLaVA-OV with SigLIP-400M
    produces 729 vision tokens for each 384 $\times$ 384 image patch, FastVLM
    outputs only 36 tokens for the same input size.

    To test whether \name{} remains effective when the vision encoder already produces
    a small number of vision tokens, we replace the SigLIP-400M encoder in LLaVA-OV
    with the FastVLM encoder. We first train a LLaVA-OV 0.5B model using the
    FastVLM vision encoder, following the original three-stage recipe on 7.1M
    samples. Then, we train \name{} on top of this model to further reduce FLOPs.
    As shown in Table~\ref{tab:fastvlm}, \name{} improves over LLaVA-OV with
    FastVLM on both Easy and Hard benchmarks while achieving 3× additional LLM
    FLOPs savings (60$\times$ vs. 20$\times$). Compared to SigLIP-based LLaVA-OV,
    the FastVLM version achieves comparable results on Easy benchmarks, but the
    gap is larger on Hard ones, confirming that aggressive token reduction mainly
    hurts performance on fine-grained benchmarks.

    \section{Re-implementation of baselines}
    In this section, we detail our re-implementation of the baselines we compare
    with, along with the hyperparameters used. Since LLaVA-OV uses a SigLIP-400M
    vision encoder that does not have a CLS token, for all methods that rely on the
    CLS token’s attention scores to select important tokens, we instead use the
    average attention each token receives from all other tokens in the sequence,
    as proposed by \citet{yang2025visionzip}.

    \textbf{VisionZip.} VisionZip~\citep{yang2025visionzip} is a token reduction
    method that selects the most important tokens, named \textit{dominant tokens},
    using the visual encoder’s attention scores. To avoid losing information, the
    remaining tokens are merged into \textit{contextual tokens} based on
    semantic similarity. In our experiments, we adapt the official VisionZip
    LLaVA-Next~\citep{liu2024llavanext} code to LLaVA-OV. Unlike LLaVA-Next, the
    LLaVA-OV image processing AnyRes strategy applies bilinear interpolation when
    the number of tokens exceeds a threshold. To prevent errors from
    interpolating removed tokens and to stay close to the original design, we remove
    this step. \citet{yang2025visionzip} also introduce VisionZip$^{\dagger}$, a
    trained version that fine-tunes the cross-modality projector. For a fair
    comparison, we train VisionZip$^{\dagger}$ on the LLaVA-OV Single-Image 3.2M
    dataset~\citep{li2024llavaone}. In all VisionZip and VisionZip$^{\dagger}$
    experiments, we set the number of retained tokens per patch to 128, split
    into 104 \textit{dominant tokens} and 24 \textit{contextual tokens}.

    \textbf{VisPruner.} VisPruner~\citep{zhang2025beyond} is a training-free
    token pruning method that retains visual tokens based on visual attention
    scores. It first selects the \textit{important tokens}, i.e., those with the
    highest scores, and then removes duplicates by keeping only \textit{diverse
    tokens} based on their similarity. As with VisionZip, we adapt the official VisPruner
    LLaVA-Next code to the LLaVA-OV backbone. In our experiments, we retain 128 tokens
    per patch, split into 96 \textit{important tokens} and 32 \textit{diverse
    tokens}.

    \textbf{HiRed.} High-Resolution Early Dropping (HiRed)~\cite{arif2025hired}
    is a plug-and-play token-reduction method designed to work under a fixed
    budget. It targets high-resolution LVLMs (i.e., LLaVA-Next) and drops tokens
    before they reach the LLM. The key idea is to evaluate the visual content of
    each image patch using the attention scores of the full image, then assign a
    budget to each patch accordingly. Within each patch, the most informative tokens
    are kept and passed to the LLM. In our experiments, we adapt the official code
    to LLaVA-OV, following the same approach used for VisionZip and VisPruner. In
    our experiments, we use the same hyperparameters as the original
    implementation, setting a token budget of 20\%.

    \textbf{M$^{3}$}. Matryoshka Multimodal Models (M$^{3}$)~\citep{cai2025matryoshka}
    represent visual content as a nested set of tokens capturing information at different
    levels of detail, from coarse to fine. The visual tokens from the encoder are
    grouped into several coarse-to-fine levels, where the coarser tokens
    $X_{S_{i-1}}$ are obtained from the finer tokens $X_{S_i}$ using average
    pooling. M$^{3}$ does not add any extra parameters. For a fair comparison,
    we train M$^{3}$ on the LLaVA-OV Single-Image 3.2M dataset~\citep{li2024llavaone},
    updating both the vision encoder and the LLM weights. In our experiments, we
    define a set of scales $\{X_{S_{i}}\}_{i=1}^{M}$ that reduce the number of
    visual tokens by factors of 1, 4, 8, and 16. For a fair comparison with our method,
    we report the results at an 8$\times$ reduction.

    \textbf{PyramidDrop}. PyramidDrop~\citep{xing2025pyramiddrop} is a
    progressive token pruning method that gradually reduces the number of visual
    tokens as the LLM depth increases. Specifically, the LLM layers are split
    into stages, and at the beginning of each stage, the number of visual tokens
    is reduced based on a predefined set of reduction rates, which are defined on
    a per-stage basis. At each pruning step, the input features are split into visual
    and text features, and then from the text features, only the features corresponding
    to the position of the last token of the user's query or instruction are kept,
    resulting in $N_{v}\times d$ visual features and a single text feature
    vector. Then, the next attention layer to be executed is applied over these
    features, with the text features as the query, resulting in image-text
    attention weights that are interpreted as a per-visual token importance
    score. These scores, together with a per-stage pre-defined drop-rate, are used
    to keep the top-k scoring visual tokens, with k as the target visual token
    to keep. This procedure is applied progressively throughout the LLM's depth at
    the beginning of each stage. In our implementation, for LLaVA-OV 0.5B with a
    24-layer LLM, we used 4 stages defined as $(1, 2-6, 7-12, 13-24)$ with drop-rates
    of $(1.0, 0.3, 0.2, 0.1)$ for easy datasets, and $(1-4, 5-10, 11-16, 17-24)$
    with drop-rates of $(1.0, 0.5, 0.25, 0.125)$ for hard datasets with average
    FLOPs saving of $4.2\times$. As for LLaVA-OV 1.5B with a 28-layer LLM we
    used 5 stages defined as $(1, 2, 3-6, 5-10, 11-28)$ with drop-rates of
    $(1.0, 0.5, 0.3, 0.2 , 0.1)$ for easy datasets, and $(1-2, 3-8, 8-12, 13-18,
    19-28)$ with drop-rates of $(1.0, 0.75, 0.5, 0.25, 0.125)$ for hard datasets
    with an average FLOPs saving of $4.6\times$.

    \textbf{SparseVLM.} Similar to PyramidDrop, SparseVLM~\citep{zhang2024sparsevlm}
    leverages the self-attention maps in the VLLM to identify the text tokens
    that are most relevant to the image, and uses them as \textit{raters} to
    score the importance of visual tokens. SparseVLM then adaptively determines how
    many visual tokens to prune at each layer based on the rank of the text-to-vision
    attention matrix, and further reduces information loss with a \textit{token
    recycling} step that aggregates the most informative pruned tokens into a smaller
    set of reconstructed tokens. It is a plug-and-play method that does not
    require additional parameters or fine-tuning.

    \section{Additional details on Centered Kernel Alignment (CKA)}

    In Figure 3 in the main manuscript, we showed how vision features evolve
    across layers within the LLM transformer of the LVLM by computing the
    Centered Kernel Alignment (CKA)~\citep{cortes2012algorithms}. More
    specifically, to compute it, let $X \in \mathbb{R}^{n \times d}$ and
    $Y \in \mathbb{R}^{n \times d}$ represent the vision features extracted from
    two different layers, where $n$ is the number of tokens and $d$ is the feature
    dimension. The CKA computation begins by forming the Gram matrices
    $K = XX^{T}$ and $L = YY^{T}$. These matrices are then centered using the centering
    matrix $H = I_{n}- \frac{1}{n}1_{n}$, where $I_{n}$ is the identity matrix
    and $1_{n}$ is an $n \times n$ matrix of ones. The centered Gram matrices are
    given by $\tilde{K}= HKH$ and $\tilde{L}= HLH$. The CKA similarity between
    $\tilde{K}$ and $\tilde{L}$ can then be computed as:
    \begin{equation}
        \text{CKA}(\tilde{K}, \tilde{L}) = \frac{\langle \tilde{K}, \tilde{L}\rangle_{F}}{\|
        \tilde{K}\|_{F}\| \tilde{L}\|_{F}}, \label{eq:cka}
    \end{equation}
    where $\langle \cdot, \cdot \rangle_{F}$ denotes the Frobenius inner product,
    and $\| \cdot \|_{F}$ represents the Frobenius norm.

\end{document}